\PassOptionsToPackage{prologue,table}{xcolor}
\documentclass[sigconf,anonymous=False]{acmart}
\renewcommand\footnotetextcopyrightpermission[1]{} 
\usepackage{tikz}
\usepackage{arydshln}
\usepackage{algorithm}
\usepackage{algpseudocode}
\algrenewcommand\textproc{\text}
\usepackage{makecell}
\usepackage{booktabs}
\usepackage{color}
\usepackage{multirow}
\usepackage{enumitem}
\usepackage{balance}
\usepackage{makecell}
\usepackage{threeparttable}
\usepackage{amsmath,amsfonts,mathtools}
\usepackage{pifont}

\usepackage{mathrsfs}
\usepackage{float}
\usepackage{graphicx}
\usepackage{subfigure}
\usepackage{xcolor}
\usepackage{enumitem}
\usepackage{hyperref}
\usepackage{tabularx, booktabs}

\newcommand{\greenfont}[1]{{\textcolor{green}{#1}}}

\AtBeginDocument{%
  \providecommand\BibTeX{{%
    \normalfont B\kern-0.5em{\scshape i\kern-0.25em b}\kern-0.8em\TeX}}}

\copyrightyear{2024}
\acmYear{2024}
\setcopyright{rightsretained}
\acmConference[KDD '24]{the 30th ACM SIGKDD Conference on Knowledge Discovery and Data Mining}{August 25--29, 2024}{Barcelona, Spain}

\newcommand{\M}{\texttt{DBLM}}

\begin{document}

\title{Time Series Supplier Allocation via Deep Black-Litterman Model}







\author{Jiayuan Luo}\authornote{Equal contribution}
\affiliation{%
  \institution{
  No Affiliation}  \city{Beijing}
  \country{China}}
\email{joyingluo@foxmail.com}

\author{Wentao Zhang\textsuperscript{*}, Yuchen Fang}
\affiliation{
  \institution{
  Wuxi Wisdom Shenshi Technology Co., Ltd, Wuxi, China}  \city{}
  \country{}}
\email{{fenqing1126,fyclmiss}@gmail.com}

\author{Xiaowei Gao}\authornote{Xiaowei Gao is the corresponding author.}
\affiliation{%
\institution{University College London}
  \city{London}
  \country{United Kingdom}}
\email{xiaowei.gao.20@ucl.ac.uk}

\author{Dingyi Zhuang}
\affiliation{
\institution{Massachusetts Institute of Technology}
\city{Cambridge}
\state{Massachusetts}
\country{USA}
}
\email{dingyi@mit.edu}

\author{Hao Chen}
\affiliation{%
  \institution{
  University of Chinese Academy of Sciences, Beijing, China}
  \city{}
  \country{}}
\email{chenhao915@mails.ucas.ac.cn}

\author{Xinke Jiang}
\affiliation{
\institution{
Peking University
}
\city{Beijing}
\country{China}
}
\email{thinkerjiang@foxmail.com}

\renewcommand{\shortauthors}{Jiayuan, Wentao et al.}

\begin{abstract}
Time Series Supplier Allocation (TSSA) poses a complex NP-hard challenge, aimed at refining future order dispatching strategies to satisfy order demands with maximum supply efficiency fully. 
Traditionally derived from financial portfolio management, the Black-Litterman (BL) model offers a new perspective for the TSSA scenario by balancing expected returns against insufficient supply risks. However, its application within TSSA is constrained by the reliance on manually constructed perspective matrices and spatio-temporal market dynamics, coupled with the absence of supervisory signals and data unreliability inherent to supplier information.
To solve these limitations, we introduce the pioneering \textbf{\underline{D}}eep \textbf{\underline{B}}lack-\textbf{\underline{L}}itterman \textbf{\underline{M}}odel (\M), which innovatively adapts the BL model from financial roots to supply chain context. Leveraging the Spatio-Temporal Graph Neural Networks (STGNNS), \M~ automatically generates future perspective matrices for TSSA, by integrating spatio-temporal dependency. 
Moreover, a novel Spearman rank correlation distinctively supervises our approach to address the lack of supervisory signals, specifically designed to navigate through the complexities of supplier risks and interactions.
This is further enhanced by a masking mechanism aimed at counteracting the biases from unreliable data, thereby improving the model’s precision and reliability. Extensive experimentation on two datasets unequivocally demonstrates \M's~ enhanced performance in TSSA, setting new standards for the field. Our findings and methodology are made available for community access and further development \footnote{\url{https://github.com/WentaoZhang2001/DBLM}}.
\end{abstract}

\vspace{-0.2cm}

\vspace{-0.15cm}

\maketitle

\vspace{-0.2cm}
\section{Introduction}

Time Series Supplier Allocation (TSSA) is a critical challenge in enhancing supply chain efficiency, characterized by its NP-hard complexity~\cite{nasiri2018incorporating,kawtummachai2005order,jayaraman1999supplier}. 
{The goal of TSSA is to develop strategies that optimize supplier capabilities and precisely match order quantities in the future, reducing discrepancies and boosting efficiency~\cite{Wan_Rao_Dong_2023,Alikhani_Torabi_Altay_2019,Gören_2018}}. 
{This task mirrors the equilibrium-return dilemma in financial management~\cite{capm1, capm2, capm3}, aiming to maximize profits while minimizing risk. {Such an analogy highlights the strategic importance of aligning supply with demand to improve overall supply chain performance.}}

In the realm of financial management, the Markowitz model has been pivotal in optimizing asset returns against portfolio risks, underpinned by the assumption of controllable profit variance~\cite{Fabozzi_Markowitz_Gupta_2015,Steinbach_2001,Sharpe1963A}. Building on this, the Black-Litterman (BL) model enhances portfolio optimization by integrating subjective investor insights as perspective matrices, thus generating portfolios that more accurately reflect expert predictions of expected returns across different investments~\cite{Black_Litterman_1991,Xing_Cambria_Malandri_Vercellis_2018}. However, the application of the BL model within supply chain management faces distinct challenges. The manual creation of perspective matrices introduces uncertainty, struggling to capture real-time market conditions and the complex, non-linear dynamics of supplier-enterprise relationships~\cite{kara2019hybrid,islam2021machine}. A notable example of this complexity is the difficulty in determining a supplier's maximum supply capabilities at a specific moment ~\cite{Alikhani_Torabi_Altay_2019,lee2009fuzzy}, such as supplier \texttt{A}'s capacity at time $t_{i-1}$, highlighted in Figure~\ref{fig:intro.png}. 

From a practical standpoint, the quest for effective TSSA in supply chains marks a promising avenue for research ~\cite{tan1998supply,chen2006fuzzy}. On the scientific front, the challenge extends beyond merely incrementally applying conventional BL models. It involves developing an effective model capable of generating comprehensive perspective matrices that accurately {reflect} supplier performance and dynamics among the supply-order gaps. Deep learning (DL) frameworks have emerged as leading solutions to capture non-linear correlations for {analysis} ~\cite{Chai_Ngai_2020,lecun2015deep}. However, pioneering within the DL framework still presents three significant challenges in this specific domain:

\begin{figure}[htbp]
  \centering
  \hspace{0cm}
\includegraphics[width=0.455\textwidth]{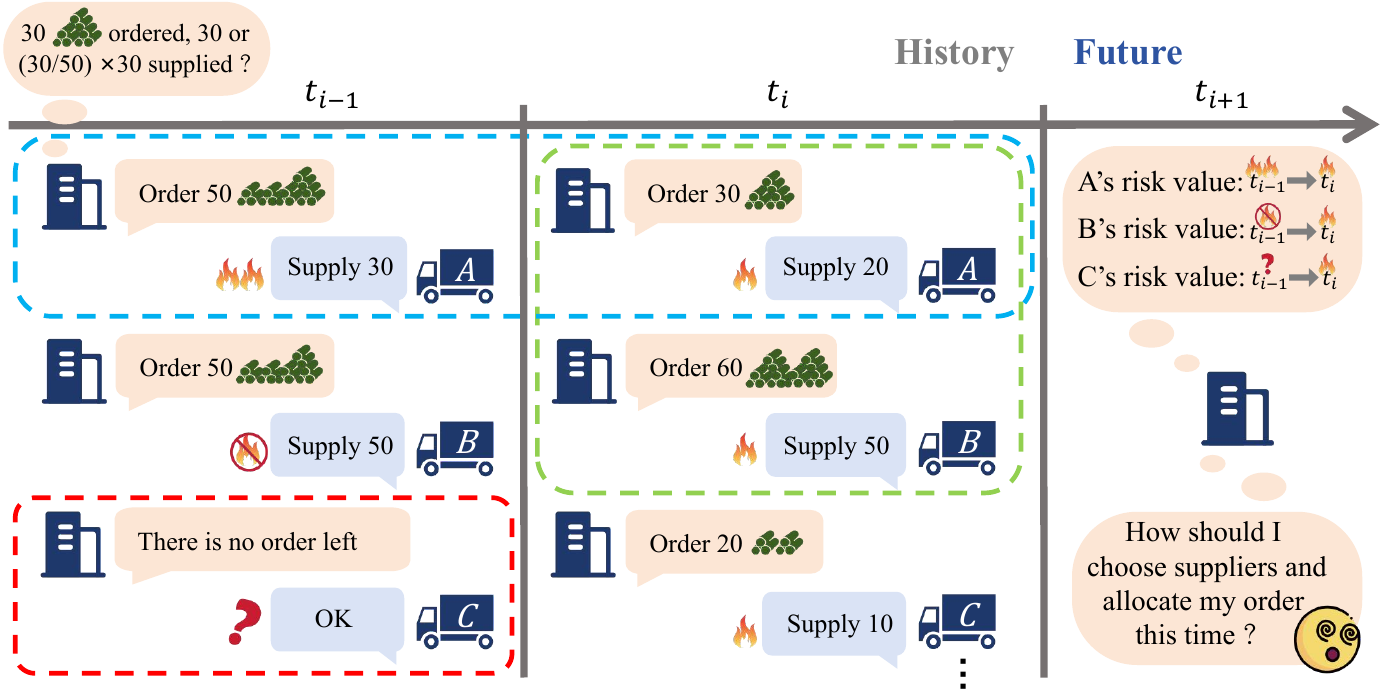}
\vspace{-0.34cm}
  \caption{An example of TSSA problem. The building symbolizes the enterprise, and the truck represents suppliers. Green signs indicate materials, while fire denotes the risk of shortfalls. The combination of a prohibition and fire signifies no risk, and a question mark indicates unknown risks.}
  \vspace{-0.2cm}
  \label{fig:intro.png}
\end{figure}

\noindent\textit{\textbf{C1. Sptaio-Temporal Dynamics.}} Supplier capacity exhibits inherent spatio-temporal dynamics crucial for future allocation. As depicted in Figure~\ref{fig:intro.png}, at time $t_i$, the allocation of orders and supplies for suppliers \texttt{A} and \texttt{B} within the green rectangle shows contrasting trends: an increase for \texttt{A} whereas a decrease for \texttt{B} compared to their previous levels at time $t_{i-1}$ respectively. These temporal trends indicate market-driven spatial competition and dynamics, which are essential for modeling future supplier capabilities. Yet, influenced by changing supplier and market inherence, these dynamics evolve over time, exemplified by \texttt{A}'s diminished supply capability from $t_{i-1}$ to $t_i$ in the blue rectangle in Figure~\ref{fig:intro.png}, highlighting the limitation of existing BL models to capture such changing spatio-temporal correlations, thus leading to inaccuracies in projecting future allocations. 

\noindent\textit{\textbf{C2. Lack of Supervisory Signals.}} Training DL models require robust supervisory signals to navigate the gradient towards the optimal direction to the objection \citep{ren2022better}. Yet, crafting these signals for comprehensive perspective matrices poses a major challenge, as accurate ground truths are hard and {unavailable} to establish, and supplementary data like supplier orders fall short in facilitating model training. Thus, our primary focus is on creating appropriate supervisory signals for perspective matrix learning.

\noindent\textit{\textbf{C3. Data Unreliability.}} Data unreliability, influenced by various factors, is a common issue in supply chain datasets, leading to biases in DL models, especially towards untraded or new suppliers with unassessed capabilities ~\cite{hao2022three,brockmann2022supply}. This is evident when supplier data, such as for untraded engaged suppliers, lacks historical engagement data, making their capabilities and risks uncertain. For example, as shown in the red rectangle in Figure~\ref{fig:intro.png}, the absence of historical orders for supplier \texttt{C} obscures their supply potential and associated risks. Hence, integrating unreliable data for spatio-temporal analysis without a robust mechanism will introduce task-irrelevant biases, undermining the effectiveness of TSSA.

In light of the aforementioned technical limitations, we propose a novel \underline{\textbf{D}}eep \underline{\textbf{B}}lack-\underline{\textbf{L}}itterman \underline{\textbf{M}}odel (\M) to provide reasonable TSSA in this paper. Specifically, the BL model is first introduced to derive the historical analytic solution of the NP-hard supplier allocation task based on the profits of suppliers and the perspectives of enterprises. In our \M~ paradigm, the perspective matrix of enterprises is data-driven instead of handcrafted, i.e., the perspective matrix in our setting is learned by training a DL to mine historical supplier capacity and market dynamics. 
Inspired by recent spatio-temporal and sequential modeling methods~\cite{fang2023spatio, SPGCL, 10.1145/3583780.3615215}, we combine the temporal and graph convolution network as the spatio-temporal graph neural networks (STGNNs) to learn the inherent spatio-temporal dynamics of historical supplier features and thus derive the perspective matrix with comprehensive knowledge to solve \textit{\textbf{C1}}. Therefore, the future solution of the supplier allocation, i.e., the goal of our task, can be approximated through the spatio-temporal information-enhanced historical one.
To introduce effective signals to address \textit{\textbf{C2}}, namely supervising the learning of future perspective matrices, we introduce a novel approach utilizing a Spearman correlation coefficient-based ranking loss. This loss function is designed around the negatively correlated weights and risks of suppliers, in line with that suppliers bearing lower risk should be awarded a greater proportion of orders.
Moreover, we design a masking mechanism during calculating the ranking loss to mitigate the bias introduced by unreliable data to address \textit{\textbf{C3}}.

In summary, our paper contributes to the field in following ways:
\begin{itemize}[leftmargin=*]
\item To the best of our knowledge, {our \M~ is the first initiative to integrate financial investment management strategies with supply chain demand challenges}. We advanced the BL model to the TSSA task, pioneering the exploration of optimal solutions and modeling the perspective matrix through a spatio-temporal graph neural network.
\item We introduce a novel masked ranking loss to guide the training of \M~, which is implemented by the Spearman rank correlation coefficient. Additionally, we implement a masking mechanism for the learned risks of unengaged suppliers within this loss function, aiming to reduce bias from unreliable data.
\item Our comprehensive experimental evaluation on two supplier allocation datasets demonstrates the superior performance of \M~ over existing baseline approaches, underscoring the efficacy and innovation of our proposed model.
\end{itemize}

\section{Related Work}

\paragraph{\textbf{Supplier Allocation}}
Traditional supplier allocation methods include: Analytic Hierarchy Process (AHP)~\cite{hamdan2017two}, Analytic Network Process (ANP)~\cite{Bayazit_2006}, Case-Based Reasoning (CBR)~\cite{Choy_Lee_Lau_Choy_2005}, Data Envelopment Analysis (DEA)~\cite{dotoli2015integrated}, Fuzzy Set Theory~\cite{Sarkar_Mohapatra_2006}, Mathematical Programming~\cite{noori2019analyzing}, and their hybrid methods\cite{moheb2019sustainable}. However, their capability in handling dynamic and uncertain issues is limited, and they demand high quality and quantity of data. For example, the AHP method requires decision-makers to compare elements at the same level based on their experience and knowledge. When new characteristic factors arise, decision-makers must review and revise their decisions~\cite{Liao_Rittscher_2007}. Methods based on fuzzy sets often fail to fully address randomness and ambiguity when solving the problem of data unclarity, and they tend to overlook the correlation of data, leading to inaccurate results~\cite{Ho_Xu_Dey_2010,islam2021machine}. ~\citet{chaising2017application} realized that in a changing organizational environment, as supplier allocation criteria may vary over time, it becomes challenging to solve the performance measurement problems of supplier allocation in the long term. They used the Fuzzy-AHP-TOPSIS method to predict and rank the optimal suppliers for different periods. ~\citet{Duică2018Selecting} employed time series analysis and linear regression analysis to predict sales performance and thereby assess and select suppliers. ~\citet{Wan_Rao_Dong_2023} developed a time-series-based multicriteria Large-Scale Group Decision-Making (LSGDM) method with intuitionistic fuzzy information to address the time-series supplier allocation issue. Despite these methods having achieved significant success in solving time series supplier allocation, they neglect the substantial influence of decision-maker's subjective factors.

\paragraph{\textbf{Black-Litterman Model}}
The Black-Litterman model is an asset allocation technique that integrates investors' market views with market benchmarks. It has become a significant improvement over Markowitz's traditional mean-variance optimization method~\cite{Black_Litterman_1991}. However, the traditional BL model often faces challenges regarding the construction of views. ~\citet{Kolm_Ritter_Simonian_2021} discussed the Bayesian interpretation of the BL model and the potential of machine learning methods to generate data-driven investment views. ~\citet{bertsimas2012inverse} expanded the BL model to encompass a more general concept of risk, thus more systematically defining the confidence levels of views. We found that the BL model's framework is not only applicable to solving portfolio problems but can also be applied to time-series supplier allocation issues. The perspective matrices built based on the BL model represent the decision-maker's assessment of different suppliers during the decision-making process, essentially translating to the allocation weight.

\paragraph{\textbf{Spatio-temporal Optimization}}
Spatio-temporal optimization is an optimization technique that combines temporal and spatial dimensions. It involves decision-making in both time and space, with the primary goal of finding the optimal or near-optimal solutions within given spatio-temporal constraints. ~\citet{Chen_Xu_Han_Fu_Pi_Joe-Wong_Li_Zhang_Noh_Zhang_2020} considered distance and separation as two dimensions in spatio-temporal optimization, creating a prediction-based actuation system for city-scale ridesharing vehicular mobile crowd-sensing. In optimizing delivery and pickup routes, ~\citet{Wen_Lin_Mao_Wu_Zhao_Wang_Zheng_Wu_Hu_Wan_2022} accounted for the impacts of different time steps on tasks and the spatio-temporal correlation between different tasks, proposing a dynamic STGNNs. ~\citet{Hui_Fang_Xia_Aykent_Ku_2023} used a structure-aware attention model to achieve Constrained Market Share Maximization under limited corporate budgets. In addressing the last-mile routing problem in logistics, ~\citet{Lyu_Wang_Song_Liu_He_Zhang_2023} considered the effects of spatio-temporal relationships on tasks and utilized a variational graph GRU encoder network to optimize the allocation of couriers in emergency last-mile logistics. These applications of spatio-temporal optimization demonstrate its advantages in solving multifaceted and interrelated tasks. However, few research applying STGNNs to optimize TSSA.

\section{Preliminaries}
\begin{definition}[Order-Supply Mechanism Data]
Consider an enterprise that consistently requires $M$ volumes of raw materials, sourced from a set of suppliers denoted as $\mathcal{SU}=\{\mathcal{SU}_1,\mathcal{SU}_2, \cdots, \\\mathcal{SU}_N\}$, where $N$ represents the total number of suppliers. This study examines the historical order demands and supply transactions between the enterprise and its suppliers over $T$ periods. The order volumes are represented by  $\mathcal{O} \in \mathbb{R}^{N\times T}$, where $\mathcal{O}_{it}$ specifies the order volume placed with supplier $\mathcal{SU}_i$ at time $t$. Similarly, supply volumes are denoted as $\mathcal{S} \in \mathbb{R}^{N\times T}$, with $\mathcal{S}_{it}$ indicating the supplied volume. Each unit volume sourced from supplier $\mathcal{SU}_i$ yields a constant return of $\mu_i (\mu_i \ge 0)$. However, supply volumes are inherently limited by the corresponding order volumes, i.e., $\forall \mathcal{SU}_i\in \mathcal{SU}, \mathcal{S}_{it}\leq \mathcal{O}_{it}$.
This constraint leads to a "Shortage of Supply" (SoS) dilemma, which significantly impacts the continuity of production and sales operations. Consequently, the enterprise is motivated to enhance its allocation strategies to meet the $M$ volume requirement efficiently by perfectly matching the supply capacity of each individual supplier and also minimizing SoS risks, by analyzing historical order and supply patterns across the market.
\end{definition}

\begin{definition}[Black-Litter Model for Supplier Allocation]
The supplier allocation challenge is conceptualized as managing a supplier portfolio, aimed at ensuring that the cumulative allocations across all suppliers in the portfolio meet the enterprise's total material requirement, $M$, while simultaneously minimizing the risk of SoS. The strategy specifies the allocation proportion or volume of orders $\mathcal{B}_i$ designated to each supplier $\mathcal{SU}_i$ within the portfolio for a given time period $t$. Importantly, these allocations are dynamic, and subject to adjustments in response to supplier performance, market conditions, and the changing needs of the enterprise. As such, the optimization task at any time $t$ aims to:
\begin{align}
\footnotesize
    \label{eq: origional_optimization11}
    \begin{cases}
        \max  & \underbrace{\sum_{i=1}^{N} \mathcal{B}_{it} \mu_i}_{\text{Profit Item}} - \delta \underbrace{\sum_{i=1}^{N} \mathcal{B}_{it}^2 (\mathcal{O}_{it} - \mathcal{S}_{it})^{\kappa}}_{\text{Risk Item}}  \\ 
        \text{subject to} &\sum_{i=1}^{N} \mathcal{B}_{it} = M, i=1,2,\cdots,N. \quad \mathcal{B}_{it} \geq 0
    \end{cases},
\end{align}
where term $\sum_{i=1}^{N} \mathcal{B}{it}^2 (\mathcal{O}{it} - \mathcal{S}{it})^{\kappa}$ quantifies the risk, with $\kappa$ being a positive integer exponent and $\delta$ as the coefficient balancing the profit and risk terms. This formula can be equivalently transformed, according to Appendix \ref{AP: Optimization Transfer}, in alignment with the BL formula, as:
\begin{align}
\footnotesize
    \label{eq: origional_optimization}
    \begin{cases}
        \max  & \underbrace{\vphantom{\mathcal{W}_{t}^T (\mathcal{O}_{t} - \mathcal{S}_{t})^{\kappa} \mathcal{W}_{t}}{\mathcal{W}_{t} \mu}}_{\text{Profit Item}} - \frac{\delta}{2} \underbrace{ \mathcal{W}_{t}^T (\mathcal{O}_{t} - \mathcal{S}_{t})^{\kappa} \mathcal{W}_{t}}_{\text{Risk Item}}  \\ 
        \text{subject to} &\sum_{i=1}^{N} {w}_{it} = 1, i=1,2,\cdots,N.\quad  1 \ge {w}_{it} \geq 0
    \end{cases},
\end{align}
where $\mathcal{W}_{t}$ is the weight allocation matrix with entry $w_{it}=\mathcal{B}_{it}/M$ signifies the proportion of total material volume allocated to supplier $\mathcal{SU}_i$ at time $t$. Here we denote diagonal matrix $\Sigma_t = (\mathcal{O}_t-\mathcal{S}_t)^\kappa$, thus the optimist weight allocation matrix is solved as $\mathcal{W}_{t}^{*}=(\delta \Sigma)^{-1}\mu$, facilitating a strategic balance between maximizing profit, as demand here, and minimizing risk.
However, notice that this optimization can only based on the existence of $\mathcal{O,S}$, which is unachievable and not suitable for future projection. As a consequence, we introduce the \M~ to adjust the parameters $\mu$ (expected returns) and $\Sigma$ (risk estimates) to better align with future conditions, effectively bridging the forecasting gap. This innovative approach allows for the recalibration of historical insights to forecast future allocation strategies more accurately, thereby enhancing the model's predictive reliability and strategic value in uncertain environments.

\end{definition}

\begin{definition}[Time Series Supplier Allocation]
Given $\{ \mathcal{ O, S,}\\ M, \mu, N \}$ observed over the historical period from $t-p$ to $t$, the objective is to determine an optimal allocation strategy for the future time interval $t+1$ to $t+f$. The main goal of \textbf{TSSA} is to optimize the future allocation weights $\mathcal{\hat{W}}^{*}\in \mathbb{R}^{N\times f}$ of all $N$ suppliers in future:
\begin{equation}
\small
    \left[\mathcal{S}_{t-p:t}, \mathcal{O}_{t-p:t}\right] \stackrel{\Theta}{\longrightarrow} \mathcal{\hat{W}}^*_{t+1:t+f},
    \label{eq:prediction objective}
\end{equation}
where $\Theta$ is the learned parameters for \M.
However, this decision-making process is complicated by the reliance on historical data pertaining only to orders $\mathcal{O}$ and supplies $\mathcal{S}$, alongside the presence of both subjective factors (e.g., the potential zeroing of order and supply data for non-engaged suppliers) and objective factors related to suppliers. Thus, achieving a globally optimal solution is impractical, rendering this an NP-hard problem. 
\end{definition}

\vspace{-0.2cm}
\begin{figure*}[t]
\vspace{-0.4cm}
  \centering
\includegraphics[width=0.9\textwidth]{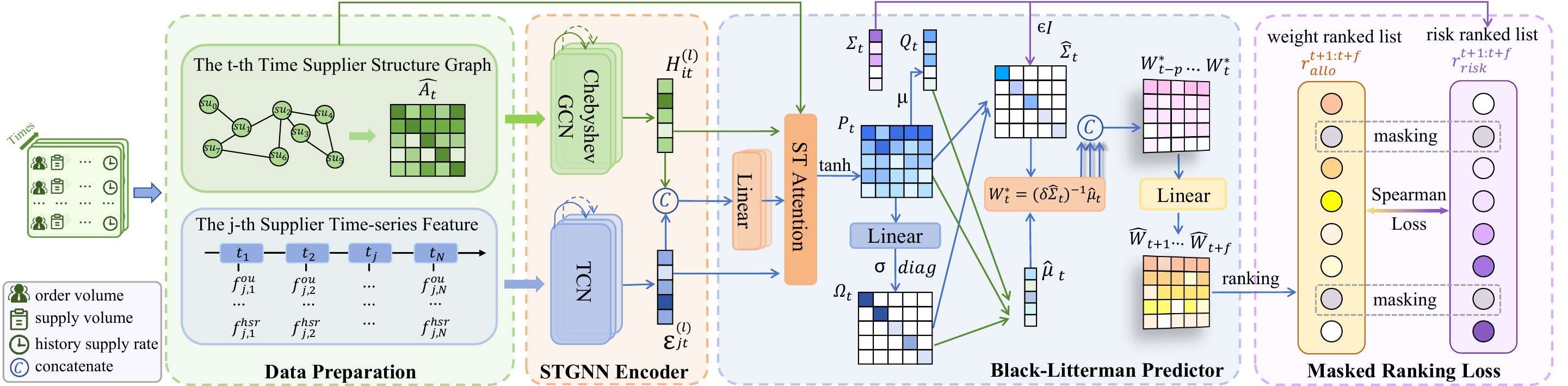}
\vspace{-0.32cm}
  \caption{Overall framework of \M. \M~constructs the feature and propagation matrix, then extracts spatial and temporal embeddings using ChebyGCN and TCN. It fuses these embeddings to obtain $\mathcal{P}$ and $\Omega$ through ST attention. Finally, \M~ solves for $\mathcal{W}^*$ in the BL model, projecting future allocations. Masked Ranking Loss is used for \M~ supervision.}
  \label{fig:method.png}
  \vspace{-0.37cm}
\end{figure*}
\section{Method}
In this section, we elaborate on our innovative \M, as illustrated in Figure \ref{fig:method.png}. Note that traditional models often struggle to incorporate the dynamic historical information of the market, leading to solutions that are optimized for past conditions but lack predictive power for future scenarios. To address this limitation, we introduce a data-driven subjective perspective from the enterprise on market dynamics. This approach allows for the adjustment of the profit term $\mu \to \hat{\mu}$ and the risk term $\Sigma \to \hat{\Sigma}$  in Section ~\ref{sec: blp}, facilitating the derivation of optimal solutions for future contexts. More specifically, we integrate the traditional BL model with advanced DL techniques by parameterizing the perspective matrix (denoted as $\mathcal{P}$) across both spatial and temporal dimensions through the STGNN encoder. Based on the principle of the BL model, we obtain the resulting adjusted solution, denoted as $\mathcal{W}^*$, and then pass it through the Predictor module to form the future allocation weight matrix $\hat{\mathcal{W}}^*$. To tackle the challenges (\textit{\textbf{C1: lack of supervisory signals}} and \textit{\textbf{C2: data unreliability}}) mentioned before, we construct a mask ranking loss.
In general, \M~ comprises the following modules:
\begin{itemize}[leftmargin=*,noitemsep,topsep=2pt]
    \item \textbf{Data Preparation} takes the supply chain data as the input, initiates the supplier feature, and constructs the dynamic structure from the market view.
    \item \textbf{STGNN Encoder} 
    employs the spatial and temporal convolution layers to learn the market's intricacies and historical patterns in low-rank representation, which are critical for parameterizing the perspective matrix to address \textbf{\textit{C1}}.
    \item \textbf{Black-Litterman Predictor}
    utilizes the spatio-temporal fusion layer to aggregate the weighted attention-based perspective matrix. Then we calculate the error covariance matrix and derive the BL solution. Given computational complexities, two specific calibrations are applied to refine the process. Moreover, for bridging the gap between the history and future optimal, we carry a predictor to translate the BL solution to future projection.
    \item \textbf{Masked Ranking Loss} is introduced to tackle \textbf{\textit{C2}} by supervising the model training with the rank approach. Furthermore, \M~ incorporates a masking strategy to circumvent issue \textbf{\textit{C3}}.
\end{itemize}
The main notations of \M~ is listed in Table \ref{tab:symbols}.
\begin{table}[!htbp]
\footnotesize
\vspace{-0.3cm}
	\centering
 \footnotesize\setlength\tabcolsep{2pt}   
	\caption{Main notations in this paper.}
 \vspace{-0.3cm}
	\begin{tabular}{c|l}
		\hline
		\rowcolor[gray]{0.9}\textbf{Notation} &  {\textbf{Description}} \\
		\hline
        $\mathcal{SU}$             & The supplier set  $\{\mathcal{SU}_1,\cdots,\mathcal{SU}_i,\cdots, \mathcal{SU}_N \}$\\
        $\mathcal{O}~/ \mathcal{S}$         & The order volume / supply volume \\
        $\mathcal{B}_i~/w_i$            & The expect supply volume / expect allocation weight for $\mathcal{SU}_i$\\
        $\Pi$          & Equilibrium Supplier Profits\\
        $\mathcal{P}~/\mathcal{Q}~/\Omega$      & Perspective/ Perspective Return / Error Covariance Matrix\\
                 $\widehat{\mu}_t$ ~/ $\widehat{\Sigma}_t$& Normalized return vector ~/ normalized risk vector at time $t$\\
         $\mathcal{W}^*$ ~/  $\hat{\mathcal{W}}^*$&  Optimal weight allocation matrix for history / future \\
        \hdashline
        $\mu_i$         & The yield per unit volume from $\mathcal{SU}_i$ \\
        $\delta$ & The reweight
coefficient of the profit item and risk item\\  
        $\kappa$   & The power value of risk item\\
        $\tau$   & The hyper-parameter to control $\mathcal{P}$ effect on the model \\
                 $\alpha_{ij,t}$ & Spatio-temporal attention between $\mathcal{SU}_i$ and $\mathcal{SU}_j$ at time $t$\\
        \hdashline
        $\mathcal{F}_t$~/$\hat{\mathcal{A}_t}$       & Feature for $\mathcal{SU}$ /Dynamic propagation matrix  at time $t$\\ 
         $\mathcal{H}_t^{(l)}$  ~/ $\mathcal{E}_t^{(l)}$        & Hidden spatial ~/ temporal representation at time $t$ of layer $l$-th \\
          $\mathbf{W}_{\text{sp}}^{(l)}$ ~/ $\mathbf{W}_{\text{te}}^{(l)}$ ~/ $\mathbf{b}_{\text{te}}^{(l)}$          & Learnable weights at layer $l$-th of spatial ~/ temporal encoder\\          $\mathbf{W}_{\text{attn}}$/$\mathbf{b}_{\text{attn}}$/ 
          $\vec{a}$ & Learnable weights at attention layer\\
         $\mathbf{W}_{\text{om}}$/
         $\mathbf{b}_{\text{om}}$         & Learnable weights  for encoding Error Covariance Matrix\\
         $\mathbf{W}_{\text{out}}/\mathbf{b}_{\text{out}}$& Learnable weights for predicting $\hat{\mathcal{W}}^*$\\
		\hline
	\end{tabular}
	\label{tab:symbols}
    \vspace{-0.5cm}
\end{table}

\vspace{-0.2cm}
\subsection{Data Preparation}
Instead of learning spatio-temporal embeddings solely from unreliable data, we extract several crucial sequence-based features of suppliers as inputs in Section~\ref{sec:prepare x}. Moreover, to capture market dynamics, we construct the spatio-temporal interactions, formed by the network of relationships among suppliers in Section~\ref{sec:prepare adj}, which satisfies the spatial input requirements of STGNNs.

\subsubsection{Feature Preparation.}
\label{sec:prepare x}
We construct eight quantitative supply indicators to serve as the initial features of the suppliers' capacity and stability.
These indicators include: \textit{supply volume} $f^{sv}$, \textit{order volume} $f^{ov}$, \textit{supply capacity} $f^{sc}$, \textit{supply stability} $f^{ss}$, \textit{supply rate} $f^{sr}$, \textit{supply shortage value} $f^{ssv}$, \textit{historical supply rate} $f^{hsr}$, and \textit{historical supply shortage value} $f^{hssv}$. The mathematical representations of these features are as follows:
\begin{equation}
\footnotesize
\begin{aligned}
    &f^{sv}_{it} = \mathcal{S}_{it}, \quad 
    f^{ov}_{it} = \mathcal{O}_{it}, \quad f^{sr}_{it} = \mathcal{O}_{it}-\mathcal{S}_{it},  \quad
    f^{ssv}_{it} = {\mathcal{S}_{it}}/{\mathcal{O}_{it}}, 
      \\
    &f^{hsr}_{it} = \frac{1}{p}\sum_{t-p}^{t}(\mathcal{O}_{it}-\mathcal{S}_{it}),  \quad
    f^{hssv}_{it} = \frac{1}{p} \sum_{t-p}^{t}\left({\mathcal{S}_{it}}/{\mathcal{O}_{it}}\right), 
      \\
    &f^{sc}_{it} = \max\{\mathcal{S}_{ij}|j\in [t-p, t]\}, 
    f^{ss}_{it} = \bigl(\sum_{j=t-p}^{t}(\mathcal{S}_{ij}-f^{sc}_{it})^2 / p\bigl)^{-{1}/{2}}. 
    \label{eq: feature transform}
    \end{aligned}
\end{equation}
Subsequently, we concatenate the indicators for $\mathcal{SU}_i$ at time $t$ into a comprehensive vector: ${f}_{it} = \left[f^{sv}_{it}, f^{ov}_{it}, f^{sr}_{it}, f^{ssv}_{it}, f^{hsr}_{it}, f^{hssv}_{it}, f^{sc}_{it}, f^{ss}_{it} \right]$. For all suppliers, we denote $\mathcal{F}_t = \{{f}_{1t}, f_{2t}, \cdots, f_{Nt}\} \in \mathbb{R}^{N \times 8}$, with ${f}_{it} \in \mathbb{R}^{8}$ as its entry. Finally, to address and mitigate issues of scalar imbalance, we apply normalization to the feature set $\mathcal{F}_t$. 

\subsubsection{Dynamic Structure Construction.}
\label{sec:prepare adj}
Furthermore, we construct a dynamic adjacency matrix to encapsulate the evolving interactions between suppliers. For each time point $t$, the Cosine similarity between suppliers $\mathcal{SU}_i$ and $\mathcal{SU}_j$ is computed as: 
$\text{sim}_t(i, j) = \frac{{f_{it} \cdot f_{jt}}}{{\|f_{it}\| \cdot \|f_{jt}\|}}$. Then derivation of the dynamic adjacency matrix $\mathcal{A}_t \in \mathbb{R}^{N\times N}$, is defined as:
$\mathcal{A}_t(i, j) = \frac{{\text{sim}_t(i, j) + 1}}{2}$,
where $\mathcal{A}_t(i, j)$ signifies the connection strength between suppliers $\mathcal{SU}_i$ and $\mathcal{SU}_j$ at time $t$. 
To minimize the impact of noise connections, redundant edges with a weight below threshold 0.3 are pruned. Finally, the normalized adjacency matrix, $\hat{\mathcal{A}_t} = \tilde{\mathcal{D}_t}^{-\frac{1}{2}}\mathcal{A}_t\tilde{\mathcal{D}_t}^{-\frac{1}{2}}$, with $\tilde{\mathcal{D}_t}$ as its degree matrix, serves as the feature propagation matrix~\cite{gpfn,katz}.

\subsection{STGNN Encoder}
In this section, we focus on encoding the prepared supplier sequence features $\{\mathcal{F}_{t-p}, \cdots, \mathcal{F}_{t} \}$ and dynamic propagation matrices $\widehat{A}_{t-p:t}$ to obtain representations in both spatial and temporal dimensions. We apply Spatial (Section~\ref{sec: scl}) and Temporal Convolution Layers (Section~\ref{sec:tcl}) to process these features and matrices, respectively. Such representations directly aid in the construction of the perspective matrix $\mathcal{P}$ (cf. Section~\ref{sec: blp}). 

\subsubsection{Spatial Convolution Layer. }
\label{sec: scl}
The spatial block of \M~ is applied to enhance supplier correlation learning by leveraging the global topological structure. In our implementation, we choose ChebGCN \citep{wu2021inductive} and the Spatial Convolution layer as:
\begin{equation}
\small
    \mathcal{H}_{t}^{(l)}  = \sigma\left(\sum_{c=1}^{C} \mathbf{T}_c (\hat{\mathcal{A}_t}) \mathcal{H}_{t}^{(l-1)} \mathbf{W}_{\text{sp}}^{(l)}\right),
\end{equation}
where $\mathcal{H}_{t}^{(l)}$ is the representation matrix in $l$-th layer at time $t$ with $\mathcal{H}_{t}^{(0)} = \mathcal{F}_t$; the Chebyshev polynomial $\mathbf{T}_c(X) = 2X\mathbf{T}_{c-1}(X) - \mathbf{T}_{c-2}(X)$ is used to approximate the convolution operation, with boundary conditions $\mathbf{T}_0(X) = I$ and $\mathbf{T}_1(X) = X$; $\mathbf{W}_{\text{sp}}^{(l)}$ is the learnable spatial convolutional kernel in layer $l$, added to control how each node transforms the received information; $\sigma(\cdot)$ is the nonlinear activation function (e.g. ReLU, Linear). In our model, we stack 3 ($C=3$) ChebGCN layers to better capture the spatial dependency.

\subsubsection{Temporal Convolution Layer. }
\label{sec:tcl}
The temporal block leverages the TCN to discern temporal trends from time-sequence features. Highlighted by \cite{wu2021spatial, zhuang2022uncertainty}, TCNs offer significant improvements over RNNs due to the ability to process input sequences of variable lengths, enhancing adaptability across different temporal scales. The general idea of TCNs lies in utilizing a shared gated 1D convolution of width $w_l$ in the $l$-th layer, facilitating the transmission of information from $w_l$ adjacent time steps. As per \citet{lea2017temporal}, each TCN layer $H_{l}$ receives the signals from the previous layer $H_{l-1}$ as:
\begin{equation}
\small
    \mathcal{E}_{t}^{(l)} = f\left(\mathbf{W}_{\text{te}}^{(l)}*\mathcal{E}_{t-1}^{(l-1)} + \mathbf{b}_{\text{te}}^{(l)}\right),
\end{equation}
where $\mathcal{E}_{t}^{(l)}$ is the representation matrix in $l$-th layer at time $t$ initialized with $\mathcal{E}_{t}^{(0)} = \mathcal{F}_t$; $\mathbf{W}_{\text{te}}^{(l)}$ is the convolution filter for the $l$-th layer, $*$ is the shared convolution operation, and $\mathbf{b}_{\text{te}}^{(l)}$ stands for the learnable bias. We also stack 3 TCN layers following \cite{zhuang2022uncertainty}.

\subsection{Black-Litterman Predictor}
\label{sec: blp}
After acquiring the spatial and temporal embeddings $\mathcal{H}_{t}^{(l)}, \mathcal{E}_{t}^{(l)}$ for each supplier, we need to base on the BL model to get the Perspective Matrix $\mathcal{P}$ and the Error Covariance Matrix $\Omega$, which are critically important in our method and are detailed in the Appendix ~\ref{ap: revisit}. We employ a spatial-temporal fusion layer to derive $\mathcal{P}$ and  $\Omega$ in Section \ref{sec:fusion}. Thus, we can bridge the gap and reconcile the historical data with future expectations by adjusting the parameters $\mu$ and $\Sigma$ with enriched spatio-temporal information. Consequently, we leverage $\mathcal{P}$ to solve and project the BL solution onto the future allocation matrix $\hat{\mathcal{W}}^*$, detailed in Section \ref{sec: solve}. 

\subsubsection{Spatio-Temporal Fusion Layer. }
\label{sec:fusion}
Upon deriving the low-rank tensors $\mathcal{H}_{t}^{(l)}, \mathcal{E}_{t}^{(l)}$, we apply an attention mechanism~\cite{fang2023spatio} to fuse spatial and temporal representations, aiming to capture the complicated dynamics of supplier interactions. Achieved through the weighted attention-based low-tensor multiplication approach, $\mathcal{P}$ learns the essential aspects of both spatial and temporal dimensions:
\begin{equation}
\small
    \begin{aligned}
    &e_{ij,t} = \text{LeakyReLU}\bigl(\vec{a}^\mathsf{T} \bigl[\mathbf{W}_{\text{attn}} [\mathcal{H}_{it}^{(l)}, \mathcal{E}_{jt}^{(l)}] + \mathbf{b}_{\text{attn}} \bigl]\bigl), \\
    &\alpha_{ij,t} = {\exp(e_{ij,t})}/{\sum_{k=1}^{N} \exp(e_{ik,t})}, \\
    &\mathcal{P}_{t} = \text{tanh}\bigl(\sum_{\hat{\mathcal{A}}_t(i, j) > 0} \alpha_{ij,t} \hat{\mathcal{A}}_t(i, j) [\mathcal{H}_{it}^{(l)} \times \mathcal{E}_{jt}^{(l)}] \bigl),
    \end{aligned}
\end{equation}
where $\alpha_{ij,t}$ is the spatio-temporal attention between $\mathcal{SU}_i$ and $\mathcal{SU}_j$ at time $t$. $\mathbf{W}_{\text{attn}}, \mathbf{b}_{\text{attn}}, \vec{a}$ are learnable matrices~\cite{GAT}. 

However, it is noted that $\mathcal{H}$ and $\mathcal{E}$ are not full-rank matrices, resulting in $\mathcal{P}$ also being non-full-rank (cf. Appendix \ref{sigmoid}). Thus, due to the data reliability, $\Sigma$ is also non-full-rank because of zero-risk non-engaged suppliers, leading to traditional computation of $\Omega_t$ falls into non-full-rank, hindering the calculation of $(\mathcal{P}\tau\Sigma\mathcal{P}^T + \Omega)^{-1}$ in BL model, as detailed in Appendix \ref{ap: onega} and Appendix \ref{ap: unj}. To circumvent this dilemma, we adopt a calibration that incorporates the non-linear Sigmoid function and utilizes deep trainable matrices to refine and enhance the learning process of $\Omega$, as detailed in Appendix \ref{ap: sigmoid_1_sol}. After sigmoid calibration, the $\Omega_t$ is learned by:
\begin{equation}
\small
\label{eq:omega}
\Omega_t = \text{diag}\bigl(\sigma(\mathbf{W}_{\text{om}}\mathcal{P}_t\Sigma_t\mathcal{P}^T_t + \mathbf{b}_{\text{om}})\bigl),
\end{equation}
where $\mathbf{W}_{\text{om}}, \mathbf{b}_{\text{om}}$ are the learnable matrices. After applying $\text{diag}(\cdot)$ and $\sigma(\cdot)$, $\Omega_t$ transfers to a full-rank matrix. Additionally, to further enhance the capability to capture and integrate spatio-temporal dynamics, we employ a multi-head mechanism for learning $\mathcal{P}_t, \Omega_t$. 

\subsubsection{Black-Litterman Solver. } 
\label{sec: solve}
The BL Model is enhanced by incorporating the perspective matrix $\mathcal{P}$ and error covariance matrix $\Omega$ from enterprise to adjust the return vector $\mu$ to $\hat{\mu}$ and risk vector $\Sigma$ to $\widehat{\Sigma}$, as detailed in Appendix \ref{ap: revisit}. Given the equilibrium supplier profits $\Pi$ normalized from $\mu$ via Eq. \eqref{eq: pi initiation}, and perspective return vector $\mathcal{Q}=\mathcal{P}\times \mu +N(0,\Omega)$, we adjust the profit and risk components as:
\begin{equation}
\small
\begin{aligned}
    \label{eq: modify} 
         \widehat{\mu}_t = ~&~ \Pi_t + \tau \Sigma_t \mathcal{P}_t^T (\mathcal{P}_t\tau\Sigma_t\mathcal{P}_t^T + \Omega_t)^{-1} (\mathcal{Q}_t-\mathcal{P}_t\Pi_t), \\
         \widehat{\Sigma}_t = ~&~ (1+\tau) (\Sigma_t+\epsilon I) - \tau (\Sigma_t+\epsilon I) \mathcal{P}_t^T \\ ~&~\times \bigl(\mathcal{P}_t \tau \Sigma_t \mathcal{P}_t^T  
         +\Omega_t\bigl)^{-1} \mathcal{P}_t\tau \Sigma_t,
\end{aligned}
\end{equation}
where $\tau$ is the hyper-parameter that regulates the impact of enterprise's perspective $\mathcal{P}_{t}$ on adjustments. Moreover, we also make regularization calibration by adding $\epsilon I$ to solve incalculaton of $\hat{\Sigma}^{-1}$ due to the non-fill-rank problem caused by data unreliability (cf. Appendix \ref{ap: rc}). By integrating $\widehat{\mu}_{t}, \widehat{\Sigma}_{t}$ into the profit and risk components, the solution:
\begin{equation}
\small
  \label{eq: BL solution2}
\mathcal{W}^*_{t} = (\delta \widehat{\Sigma}_{t})^{-1} \widehat{\mu}_{t},
\end{equation}
where $\mathcal{W}^*_{t}$ denotes the history optimal allocation strategy, ensuring $\sum \mathcal{W}^*_{t} = 1$. However, it is still struggling to adapt this historically optimal BL solution to align with future projections. Since future optimal allocation exhibits intricate relationship gaps with $\mathcal{W}^*_{t}$. 
Consequently, for mapping the optimal allocation into future intervals \cite{mlpt1, mlpt2}, a non-linear transformation is applied on the combination of previous allocations $\{\mathcal{W}_{t-p}^*, \cdots, \mathcal{W}_{t}^*\}$:
\begin{equation}
\small
\begin{aligned}
\label{eq:OUT}
\mathcal{\hat{W}}^*_{t+1:t+f} = \text{Softmax}(\mathbf{W}_{\text{out}} {\mathcal{W}}_{t-p:t}^* + \mathbf{b}_{\text{out}}),
\end{aligned}
\end{equation}
where $\mathbf{W}_{\text{out}}, \mathbf{b}_{\text{out}}$ are learnable matrices. $\text{Softmax}(\cdot)$ is applied time-wisely, to
ensure the sum of allocation weights normalized to 1.
\subsection{Loss Function and Model Training}
Due to the challenges posed by \textit{\textbf{C1: lack of supervisory signals}} and \textit{\textbf{C3: data unreliability}},  traditional loss functions like Mean Squared Error (MSE)~\cite{wang2009mean} and Cross-Entropy (CE)~\cite{de2005tutorial} tend to underperform. Consequently, we introduce a novel mask ranking loss crafted-carefully to address the two challenges in Section \ref{sec: mrl}. Moreover, we detail the training algorithm of \M~ in Section \ref{sec:training}.
\subsubsection{Masked Ranking Loss}
\label{sec: mrl}
 We denote the risk list $(O_{t}-S_t)^\kappa$ as risk ranked list $\mathbf{r}_{\text{risk}}^t$ in ascending order and predicted allocation weight $\mathcal{\hat{W}}^*_{t}$ as allocation ranked list $\mathbf{r}_{\text{allo}}^t$ in descending order. The two lists are converted to sequences of ranks, where $r_i^t$ indicates the rank of the $i$-th item at time $t$. Then, the two ranks are leveraged to compute the negative part of the Spearman correlation coefficient~\cite{mitra2014multivariate} as supervised loss $\mathcal{L}$. By minimizing $\mathcal{L}$, we can optimize the monotonic trend between two ranked sequences to address \textit{\textbf{C1}}:
\begin{equation}
\small
\begin{aligned}
\label{eq:loss}
\min_{\Theta}\mathcal{L} = \sum_{t=0}^{T_{train}} \sum_{j=t}^{t+f} 
\frac{6 \sum_{i=1}^{N-|\mathcal{S}_{j}|} (r_{i,\text{risk}}^{j} - r_{i,\text{allo}}^{j})^2}{(N-|\mathcal{S}_{j}|)\bigl((N-|\mathcal{S}_{j}|)^2 - 1\bigl)} + \eta ||\Theta||^2,
\end{aligned}
\end{equation}
where $N-|S_{j}|$ indicates we only calculate the gradients where $\mathcal{S}_{ij}\neq 0$ to solve \textit{\textbf{C3}}. Weight decay regularization $||\Theta||^2$ weighted by $\eta$ is applied to mitigate overfitting.
In the implementation, we adopt a masking mechanism, and due to the loss of gradient information after ranking, we utilize soft rank as suggested by \cite{blondel2020fast}.

\subsubsection{Model Training}
\label{sec:training}
Algorithm \ref{alg:training} outlines the comprehensive training process of \M.
Given the supply chain data $\mathcal{O}, \mathcal{S}$, \M~ commences with processing the attribute $\mathcal{F}_t$ and constructing dynamic propagation matrices $\hat{A}$, and initializing the network parameters (Lines 1-2).
Following this, the Spatial and Temporal Convolution layers are deployed to generate the representation $\mathcal{H}$ and $\mathcal{E}$ respectively (Line 4). 
These representations are then integrated to obtain a spatio-temporal embedding, from which the Perspective Matrix $\mathcal{P}$ is computed via the attention mechanism (Line 5)
Concurrently, the Error Covariance Matrix $\Omega$ is formulated from $\mathcal{P}$ (Line 6).
Then the BL solver is applied to ascertain $\mathcal{W}^*$ (Line 7) and calculate $\mathcal{\hat{W}}^*_{t+1:t+f}$ via non-linear transformation (Line 8) to bridge the gap. Finally, we calculate $\mathcal{L}$ and optimize it by the Adam optimizer with L2 regularization (Line 9). 
The whole process time complexity is $\mathcal{O}(TN^3p)$ and the details are as illustrated in Appendix \ref{ap: algorithm complexity}.
\begin{algorithm}[H]
    \caption{Training Procedure of \M.}
    \footnotesize
    \label{alg:training}\renewcommand{\algorithmicensure}{\textbf{Output:}}
    \begin{algorithmic}[1]
    \Require Suppliers set $\mathcal{SU}$, supply chain data $\mathcal{O},\mathcal{S}$, total target volume $M$, hyper-parameters $\delta, \kappa, \tau, l, \epsilon, \eta$. 
    \State Prepare feature $\mathcal{F}_t$ and construct dynamic propagation matrix $\hat{\mathcal{A}_t}$. 
    \State Initialize parameters $ \Theta=\{\mathbf{W}_{\text{sp}}^{(l)},\mathbf{W}_{\text{te}}^{(l)},\mathbf{b}_{\text{te}}^{(l)},\mathbf{W}_{\text{attn}}, \mathbf{b}_{\text{attn}},  \vec{a},
    \mathbf{W}_{\text{om}},
    \mathbf{b}_{\text{om}}, \mathbf{W}_{\text{out}},$ $\mathbf{b}_{\text{out}} \}$ via Xavier Initializer.
    \While { $\mathcal{L}$ does not converge}     \Comment{\textbf{Train}}
    \State Encode spatial $\mathcal{H}_t^{(l)}$ and temporal $\mathcal{E}_t^{(l)}$ representation via ChebGCN and TCN;
    \State Fuse $\mathcal{H}_t^{(l)}$ and $\mathcal{E}_t^{(l)}$ to construct Perspective Matrix $\mathcal{P}_t$; \Comment{Fusion}
    \State Drive Error Covarianze Matrix $\Omega_t$ by $\mathcal{P}_t$ via Eq.\eqref{eq:omega};    
    \State Calculate history optimist $\mathcal{W}^*$ via Eq.\eqref{eq: modified profit and risk item} and Eq.\eqref{eq: BL solution2} by $\mathcal{P}_t, \Omega_t$;  \Comment{BL Solve}
    \State Predict $\mathcal{\hat{W}}^*_{t+1:t+f}$ via Eq.\eqref{eq:OUT};   \Comment{\textbf{Predict}}
    \State Minimizing $\mathcal{L}$ via Eq.~\eqref{eq:loss} using Adam Optimizer;   
    \EndWhile
    \State End optimizing parameters $\Theta$;  
    \\
    \textbf{return} predicted low-risk allocation weight matrix $\mathcal{\hat{W}}^*$.
    \end{algorithmic}
\end{algorithm}

\section{Experiments}
In this section, we explore the efficacy of our proposed \M, in optimizing allocation task for time-series suppliers on two benchmark datasets. Our investigation is aimed at addressing the following pivotal research questions:
\begin{itemize}[leftmargin=*]
    \item \textbf{RQ1:} How does \M~ compare with state-of-the-art approaches in optimizing allocation SoS risk for time-series suppliers?
    \item \textbf{RQ2:} What contributions do the key components of \M~ make to improving supplier allocation outcomes?
    \item \textbf{RQ3:} Can the perspective matrix learned through STGNNs significantly enhance the optimization of allocation risk?
    \item \textbf{RQ4:} How does the model's performance vary with adjustments to the risk coefficient $\delta$ and the reweight coefficient $\eta$?
\end{itemize}
In addition to the main experiment to answer the four RQs, we also conduct robustness analysis in Appendix \ref{ap: robutness}.

\subsection{Experiment Setup}

\subsubsection{Datasets}
We evaluate \M~ on two supply chain datasets to optimize Time Series Supplier Allocation (TSSA), namely \textit{MCM} and \textit{SZ}, to ensure comprehensive validation. The \textit{MCM} dataset comprises supply and order data from 401 suppliers over 240 weeks, and the \textit{SZ} dataset includes data from 218 suppliers across 731 days.

\subsubsection{Compared Methods.}
We benchmark \M~ against 14 methods across the \textit{MCM-TSSA} and \textit{SZ-TSSA} datasets, categorized as follows:
\begin{itemize}[leftmargin=*]
    \item \textbf{Basic Algorithms:} \textbf{HA}~\cite{10.1609/aaai.v33i01.3301922} and \textbf{MC}~\cite{Puka2022FuzzyMA}.
    \item \textbf{Decision-making Algorithms:} \textbf{Greedy}~\cite{9835149},\textbf{DP}~\cite{Kuroiwa_2023}, \textbf{Fuzzy-AHP}~\cite{rezaei2020supplier}, \textbf{Fuzzy-TOPSIS}~\cite{hasan2020resilient} and \textbf{Markowitz}~\cite{way2018wright}.
    \item \textbf{Machine Learning Methods:} textbf{DT}~\cite{Bowser_Chao_1993}, \textbf{Lasso}~\cite{Tibshirani1996}, \textbf{MLP}~\cite{MLP}, 
\textbf{ECM}~\cite{Xing_Cambria_Malandri_Vercellis_2018}, \textbf{SGOMSM}~\cite{hui2023constrained} and 
\textbf{AGA}~\cite{li2021large}.
\end{itemize}
The basic and decision-making algorithms aim at identifying historically risk-optimal portfolios, whereas the machine learning methods, including our \M~model, utilize predictive analytics to assemble portfolios optimized for future allocation risks. The implementation details can be referred to Appendix \ref{ap: baselines details}.

\begin{table*}[!t]
\setlength{\tabcolsep}{1.1mm}
\vspace{-0.3cm}
\caption{Performance comparison (in mean ± standard deviation) on \textit{MCM-TSSA} and \textit{SZ-TSSA}. The best performance is shown in \textcolor{red!60}{\textbf{red cells and bolded}}, and the second runners are shown \greenfont{green cells} in baselines, \textcolor{cyan!60}{blue cells} in ablation study.}
\vspace{-0.2cm}
\footnotesize
\label{tab:comparison}
\centering
\begin{tabular}{cc|cccc|cccc}
\hline
\multirow{2}{*}{\textbf{Method}} & \multicolumn{1}{c|}{\textbf{Dataset}} & \multicolumn{4}{c|}{MCM-TSSA} & \multicolumn{4}{c}{SZ-TSSA} \\
\cline{2-10}
& \textbf{Metric} & HR@10 & HR@20 & HR@50 & MRE & HR@10 & HR@20 & HR@50 & MRE \\
\hline
\multirow{13}{*}{\textbf{Baselines}} 
& HA & 0.045$\pm$0.087 & 0.125$\pm$0.058 & 0.268$\pm$0.049 & 0.968$\pm$0.092 & 0.039$\pm$0.086 & 0.104$\pm$0.117 & 0.230$\pm$0.099 & 0.929$\pm$0.087  \\
& MC &0.053$\pm$0.096 &0.148$\pm$0.072  &0.276$\pm$0.087 &0.924$\pm$0.053 & 0.059$\pm$0.147& 0.141$\pm$0.152& 0.245$\pm$0.104  & 0.859$\pm$0.057  \\ 
\cdashline{2-10}
& Greedy &0.078$\pm$0.050 &0.166$\pm$0.061  &0.307$\pm$0.044 &0.902$\pm$0.108 &0.072$\pm$0.088 & 0.154$\pm$0.120 &0.349$\pm$0.109 &0.995$\pm$0.149 \\ 
& DP  & 0.075$\pm$0.082& 0.155$\pm$0.070& 0.303$\pm$0.053& 0.930$\pm$0.124& 0.069$\pm$0.075 &  0.137$\pm$0.096&  0.346$\pm$0.142& 0.942$\pm$0.155\\ 
& Fuzzy-AHP & 0.204$\pm$0.197 & 0.241$\pm$0.132 & 0.311$\pm$0.155 & 0.897$\pm$0.162 & 0.169$\pm$0.098 & 0.217$\pm$0.133 & 0.306$\pm$0.129 & 0.742$\pm$0.140 \\
& Fuzzy-TOPSIS & 0.104$\pm$0.128 & 0.187$\pm$0.140 & 0.233$\pm$0.165 & 0.887$\pm$0.143 &0.095$\pm$0.087 & 0.127$\pm$0.094 & 0.149$\pm
$0.138 & 0.939$\pm$0.143 \\
& Markowitz & 0.139$\pm$0.170 & 0.227$\pm$0.158 & 0.309$\pm$0.106 & 0.997$\pm$0.191 &  0.118$\pm$0.149  & 0.154$\pm$0.110 & 0.289$\pm$0.128 & 0.844$\pm$0.185 \\
\cdashline{2-10}
& DT  & 0.040$\pm$0.492& 0.098$\pm$0.524& 0.204$\pm$0.460& 0.974$\pm$0.680&0.038$\pm$0.612 & 0.106$\pm$0.598&  0.206$\pm$0.720& 0.977$\pm$0.749\\ 
& Lasso  & 0.066$\pm$0.544& 0.137$\pm$0.670& 0.296$\pm$0.399& 0.872$\pm$0.721&0.061$\pm$0.482 & 0.161$\pm$0.670&  0.350$\pm$0.648&0.736$\pm$0.725\\ 
& MLP & 0.199$\pm$0.344 & 0.245$\pm$0.287 & 0.331$\pm$0.225 & 0.973$\pm$0.339 & 0.182$\pm$0.291 & 0.246$\pm$0.348 & 0.382$\pm$0.306 &  0.556$\pm$0.320  \\ 
& ECM & \cellcolor{green!15}0.272$\pm$0.282 & 0.289$\pm$0.299 & \cellcolor{green!15}0.348$\pm$0.310 &  \cellcolor{green!15}0.641$\pm$0.407 & \cellcolor{green!15}0.253$\pm$0.238 & \cellcolor{green!15}0.290$\pm$0.288 & \cellcolor{green!15}0.412$\pm$0.271 & \cellcolor{green!15}0.493$\pm$0.377 \\
& SGOMSM & 0.263$\pm$0.397& \cellcolor{green!15}0.311$\pm$0.403&0.327$\pm$0.454 & 0.844$\pm$0.429& 0.204$\pm$0.140&0.282$\pm$0.198 &0.369$\pm$0.245 & 0.671$\pm$0.298\\ 
& AGA & 0.158$\pm$0.237 &0.206$\pm$0.228 & 0.310$\pm$0.296&0.772$\pm$0.357&0.180$\pm$0.205&0.242$\pm$0.167 &0.374$\pm$0.152 & 0.629$\pm$0.261 \\ 
\cdashline{1-10}
\textbf{Ours} & \textbf{\M} & \cellcolor{red!20}{\textbf{0.403$\pm$0.284}} & \cellcolor{red!20}{\textbf{0.449$\pm$0.293}} & \cellcolor{red!20}{\textbf{0.487$\pm$0.356}} & \cellcolor{red!20}{\textbf{0.518$\pm$0.292}} & \cellcolor{red!20}{\textbf{0.481$\pm$0.158}} & \cellcolor{red!20}{\textbf{0.543$\pm$0.187}} & \cellcolor{red!20}{\textbf{0.662$\pm$0.182}} & \cellcolor{red!20}{\textbf{0.327$\pm$0.323}}   \\
\hline
\rowcolor[gray]{0.95} \multicolumn{2}{c|}{*\textbf{Performance Gain $\uparrow$}} & 0.482  & 0.444 & 0.399 & 0.192 & 0.901 & 0.872 &  0.607& 0.337 \\ \hline
\multirow{7}{*}{\textbf{Ablation}} & \M (w/o BL) & 0.154$\pm$0.488 & 0.238$\pm$0.462 & 0.347$\pm$0.529 & 0.820$\pm$0.442 & 0.112$\pm$0.658 & 0.148$\pm$0.495 & 0.325$\pm$0.431 & 0.729$\pm$0.480 \\
& \M~(w/o STGNN) & 0.306$\pm$0.280  & 0.348$\pm$0.305 & 0.377$\pm$0.340 & 0.852$\pm$0.319 & 0.274$\pm$0.144 & 0.309$\pm$0.195 & 0.420$\pm$0.170 & 0.438$\pm$0.266 \\
& \M~(w/o TCN) & 0.314$\pm$0.277 & 0.370$\pm$0.284 & 0.393$\pm$0.342 & 0.648$\pm$0.320 &  0.293$\pm$0.109 & 0.341$\pm$0.132 & 0.448$\pm$0.240 & \cellcolor{cyan!20}{0.361$\pm$0.258}  \\
& \M~(w/o DGCN) & 0.323$\pm$0.211 & 0.419$\pm$0.277 & 0.431$\pm$0.330 & 0.719$\pm$0.376 &  0.340$\pm$0.172& \cellcolor{cyan!20}{0.442$\pm$0.249} & 0.473$\pm$0.150&0.377$\pm$0.342 \\
& \M~(w/o Fusion) & \cellcolor{cyan!20}{0.379$\pm$0.299} & \cellcolor{cyan!20}{0.420$\pm$0.311} & \cellcolor{cyan!20}{0.453$\pm$0.328} & \cellcolor{cyan!20}{0.588$\pm$0.347} & \cellcolor{cyan!20}{0.364$\pm$0.455} & 0.425$\pm$0.298 & \cellcolor{cyan!20}{0.588$\pm$0.211} & 0.367$\pm$0.243 \\
& \M~(w/o Mask) & 0.376$\pm$0.280 & 0.390$\pm$0.246 & 0.426$\pm$0.359 & 0.626$\pm$0.341 & 0.349$\pm$0.240 & 0.426$\pm$0.328 & 0.539$\pm$0.331 & 0.427$\pm$0.397 \\
& \M~(w/o Rank Loss) & 0.290$\pm$0.279 & 0.317$\pm$0.194 & 0.335$\pm$0.243 & 0.692$\pm$0.287 & 0.307$\pm$0.198 & 0.373$\pm$0.276 & 0.501$\pm$0.453 & 0.486$\pm$0.493 \\
\hline
\end{tabular}
\vspace{-0.3cm}
\end{table*}

\subsubsection{Evaluation Metrics.} 
Given the challenges highlighted in \textit{\textbf{C3}} regarding the unreliability of predicted ground-truth allocation data, traditional evaluation metrics such as Mean Absolute Error (MAE) and Root Mean Squared Error (RMSE) are deemed less suitable for our context. Therefore, we opt for two alternative evaluation metrics, \textbf{Hit Ratio@K (HR@K)} and \textbf{Mask Risk Expect (MRE)}, to more accurately and fairly evaluate the model's performance. For the details of evaluation matrics can be referred to Appendix \ref{sec: evaluation matrics}. In our implementation, we select HR@10, HR@20, HR@50, and MRE as the key metrics for evaluating our model's performance.

\subsubsection{Reproducibility.}
To ensure reproducibility, we optimize the parameters of baseline models using the Adam Optimizer with $L_2$ regularization and a dropout rate of 0.2. The sequence length for both input and forecasting is set to $p=f=4$. For the \M~ model and baselines incorporating GNNs and TCNs, we utilize three layers with 150 hidden units each.
For the \M~ model specifically, we set $\tau=3, \delta=0.6, \eta=1e^{-4}$, and $\kappa=2$, with the number of attention heads at 3 and a soft rank regularization strength of 0.5. An early-stopping strategy with a patience of 10 epochs is employed to mitigate overfitting. The dataset is divided into 70\% training, 10\% validation, and 20\% testing portions.
Implementations are done using the PyTorch 1.9.0 framework~\cite{paszke2019pytorch} in Python 3.8, on an Ubuntu server equipped with an NVIDIA Tesla V100 GPU and an Intel(R) Xeon(R) CPU. ECM, SGOMSM, and AGA are implemented using author-provided source codes, and other baselines are all implemented by ourselves.

\subsection{Main Results (RQ1)}
\label{Main Results}
To answer RQ1, we conduct experiments and report results of performance comparison on \textit{MCM} and \textit{SZ} datasets in Table 3. From the reported HR@K and MRR, we can find the following observations:

\textbf{Comparison with Traditional Methods. } Decision-making algorithms (Greedy, DP, Fuzzy-AHP, Fuzzy-TOPSIS, Markowitz) demonstrate superior performance over basic approaches (HA, MC) in terms of HR@K and MRE. This indicates that utilizing historical data can enhance future risk allocation strategies to some extent. However, these algorithms are considerably less effective than machine learning models, underscoring the critical importance of leveraging predictive analytics for addressing the TSSA challenge \textbf{\textit{C1: Spatio-Temporal Dynamics}}. This highlights the limitations of decision-making and basic methods, which rely solely on historical data. While DT, Lasso, MLP, and AGA attempt to capture temporal dependencies, their failure to adequately address the complex spatio-temporal interplay results in suboptimal outcomes, which can not reveal the variance in the supplier market.  ECM and SGOMSM, optimized for allocation tasks and utilizing STGNNs, significantly outperform other baselines but falter in managing data unreliability, treating unknown or unordered suppliers as zero or no risk, which severely includes bias and undermines their efficacy in pinpointing high-risk entities. 
In contrast, \M~ leverages a Mask Rank Loss to deal with data unreliability and incorporates a perspective matrix that spans both spatial and temporal dimensions, offering a more nuanced understanding of market dynamics, and surpassing all baselines.

\textbf{Consistent Performance Superiority. } The \M~ model consistently outperforms all baselines, with improvements ranging from 19.2\% to 48.2\% on the \textit{MCM} dataset and 33.7\% to 90.1\% on the \textit{SZ} dataset. Notably, the model's performance excels at higher HR@K values, especially for larger K values, such as HR@50. For instance, \M~ secures an impressive hit ratio of 48.7\% on \textit{MCM} and 66.2\% on \textit{SZ}, emphatically demonstrating the superiority of utilizing STGNNs to address the TSSA issue. This capability significantly empowers decision-makers to sidestep high-risk suppliers, offering critical insights from both market and temporal perspectives.

\begin{figure}[htbp]
\vspace{-0.2cm}
  \centering
  \hspace{0cm}
\includegraphics[width=0.45\textwidth]{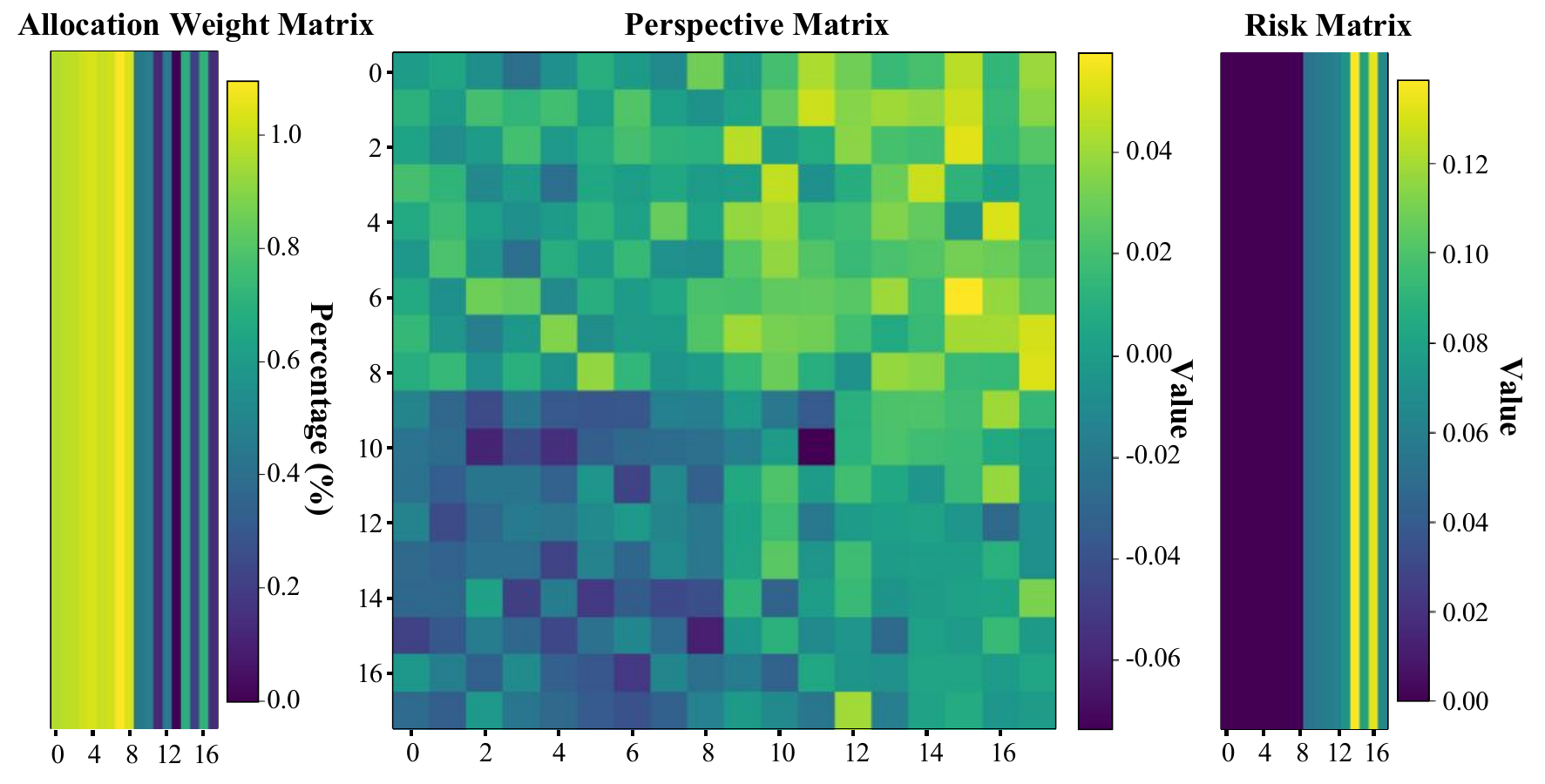}
\vspace{-0.3cm}
  \caption{ The Risk Matrix (\textbf{Left}.) composed of the top 9 and bottom 9 suppliers sorted by ascending risk, along with their corresponding Allocation Weight Matrix (\textbf{Right}.) and Perspective Matrix (\textbf{Middle}.). }
  \label{fig:hot_figures.PNG}
  \vspace{-0.5cm}
\end{figure}

\subsection{Ablation Study (RQ2)}
\label{Ablation Study}
To address RQ2, we perform ablation studies to evaluate the impact of each component within \M, summarized in Table \ref{tab:comparison}, with the following seven variants: 
\begin{enumerate}[leftmargin=*]
    \item \M~ without BL models, utilizing Markowitz weights and spatio-temporal embedding for prediction (denoted as w/o BL),
    \item \M~ without the STGNN encoder, replaced by an MLP (denoted as w/o STGNN),
    \item \M~ without the TCN encoder, relying solely on the DGCN encoder for $\mathcal{P}$ generation (denoted as w/o TCN),
    \item \M~ without the DGCN encoder, using only the TCN encoder (denoted as w/o DGCN),
    \item \M~ without Fusion Attention, opting to combine spatial and temporal embeddings directly rather than using the attention mechanism (denoted as w/o Fusion),
    \item \M~ without Mask Rank Loss, ignoring the data unreliability and removing mask mechanism (denoted as w/o Mask),
    \item \M~ without Rank Loss, using only MAE loss to verify the lack of supervisory signals (denoted as w/o Rank Loss).
\end{enumerate}

Table \ref{tab:comparison} reveals that each component makes a positive contribution to \M's overall performance. The exclusion of any part leads to a reduction in effectiveness. 
Notably, the removal of the BL component results in the most pronounced performance degradation. This underlines BL's vital role in modeling the perspective matrix within \M.
Furthermore, the absence of STGNN, TCN, or DGCN modules illustrates the necessity of capturing spatio-temporal dependencies to effectively tackle the TSSA challenge, in line with \textit{\textbf{C1: Spatio-Temporal Dynamics}}. Notably, the removal of the DGCN has a slightly less detrimental effect than the absence of the TCN module, emphasizing the critical importance of temporal dynamics in the TSSA task. Moreover, the variant without Rank Loss exhibits the second-largest decrease in performance metrics such as HR@K and MRE, following the variant without BL. This indicates that the lack of supervisory signals, \textit{\textbf{C2: Lack of Supervisory Signals}}, significantly impairs the model's effectiveness, highlighting the critical role of adopting appropriate loss functions. This rationale similarly applies to w/o Mask, providing a profound understanding of the contributions of Mask Mechanism to address \textbf{\textit{C3: Data Unreliability}} across two datasets.

\subsection{Visualization (RQ3)}
To answer RQ3, we visualize allocation weights $\hat{\mathcal{W}}$ and explain the learned perspective matrix $\mathcal{P}$ from the initial time horizon of the test set. Indeed, we specifically concentrate on distinguishing between the top and bottom 9 suppliers ranked by ground-truth risk, after initially filtering our unreliable data. We categorize suppliers numbered 0-8 as those embodying the lowest risk (top 9) and suppliers numbered 9-17 as those representing the highest risk (bottom 9), arranging them in ascending order from the lowest to highest risk values. 
Subsequently, we correlate each supplier with their respective allocation weight referencing their risk rank. The illustrations in Figure~\ref{fig:hot_figures.PNG}—specifically, its left and right subgraphs—demonstrate that the allocation weight percentages with the lowest risk, conspicuously exceed those for the higher-risk suppliers, with values nearing 1\%. This observation is in line with our theoretical assertion that suppliers bearing lower risk should be awarded a greater proportion of orders.

Moreover, we analyze the relative perspective from the decision-maker of the top and bottom 9 suppliers based on risk ranking, which is in the middle of Figure~\ref{fig:hot_figures.PNG}. Within this matrix, lighter shades of $\mathcal{P}_{ij}$ indicate supplier $\mathcal{SU}_i$ enjoys higher competitiveness compared to $\mathcal{SU}_j$, whereas darker shades signal lower competitiveness to $\mathcal{SU}_j$. 
Observations from the matrix reveal that shading near upper-left and lower-right corners, tends towards values close to 0, indicating that the differences within each group (top or bottom 9 suppliers) are considerably less pronounced than those between two groups. Whereas, lighter shades are predominantly in upper-left quadrant and darker shades in lower-right, suggesting that the top 9 suppliers, by their nature, are superior and more competitive than the bottom 9, which coincidentally corresponds to high-risk and low-risk ones. The insightful comparison 
within $\mathcal{P}$ provides empirical validation for our theoretical proposition that suppliers with lower risk are inherently more competitive in the market.

\begin{figure}[htb]
\vspace{-0.3cm}
  \centering
  \hspace{6cm}
  \includegraphics[scale=0.27]{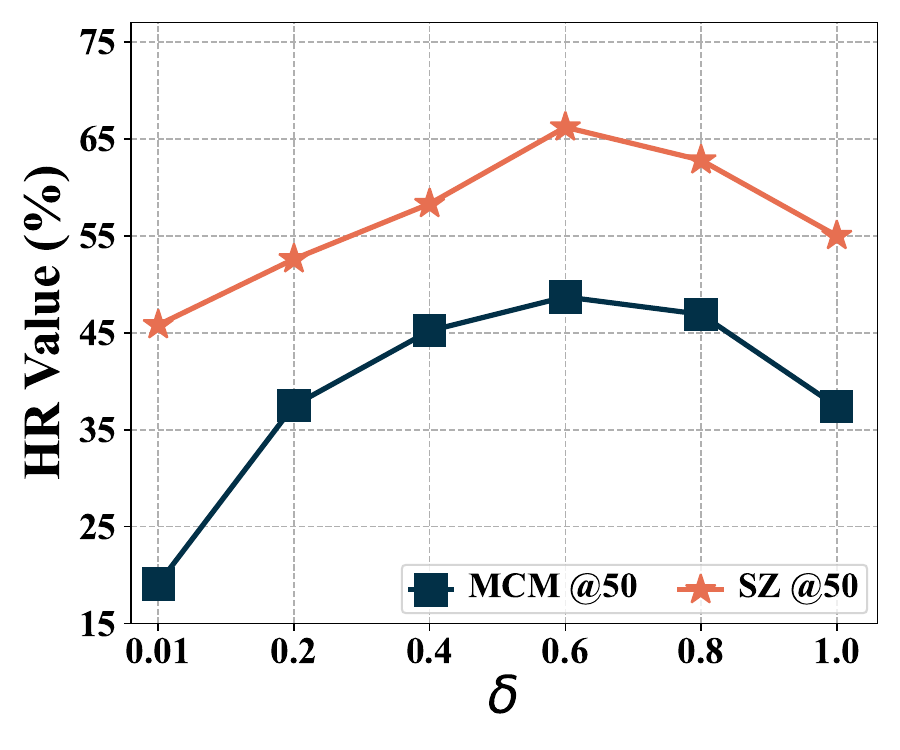}
 \includegraphics[scale=0.27]{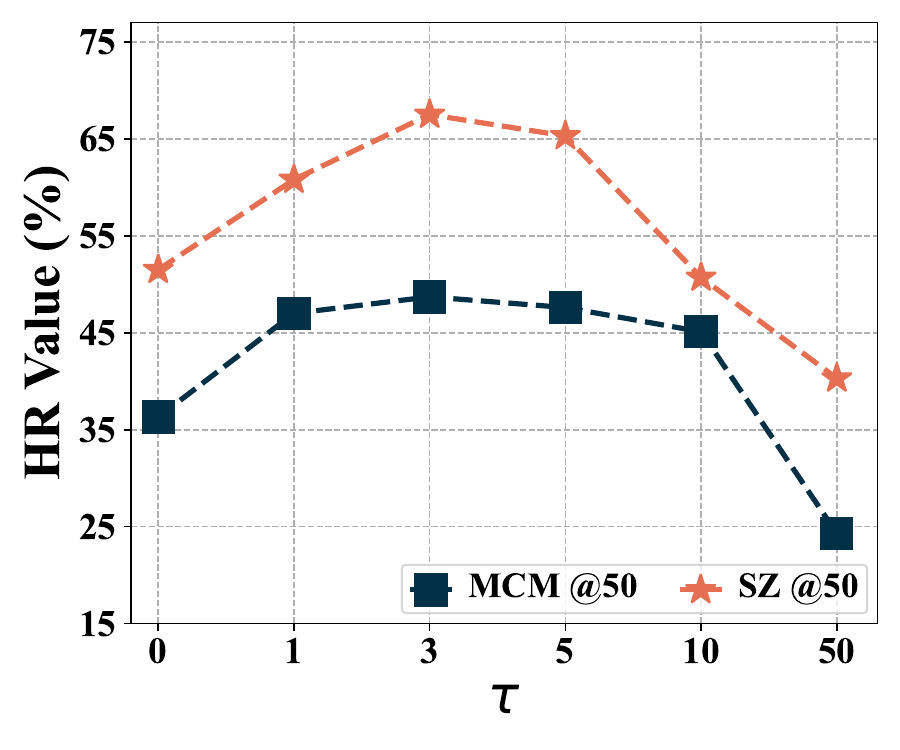}
  \vspace{-5pt}
  \vspace{-0.3cm}
  \caption{(\textbf{Left}.) Hyper-parameter study with  $\delta$ on \textit{MCM} and \textit{SZ} datasets from 0.01 to 1.0. (\textbf{Right}.) Hyper-parameter study with  $\tau$ on \textit{MCM} and \textit{SZ} datasets from 0 to 50.
  }
   \label{fig:hyper_study}
   \vspace{-0.3cm}
\end{figure}

\subsection{Hyper-parameter Study (RQ4)}
\label{Hyper-parameter Study}
This section delves into the impact of varying hyper-parameters on the performance of \M~, addressing RQ4. In Figure~\ref{fig:hyper_study}, we explore two critical hyper-parameters: the profit-risk weighting coefficient $\delta$ and the modulation coefficient $\tau$ affecting the influence of $\mathcal{P}$. The left segment of Figure~\ref{fig:hyper_study} analyzes the HR@50 performance on \textit{MCM} and \textit{SZ} datasets as $\delta$ ranges from 0 to 1, while the right segment assesses $\tau$'s effect, with its value spanning from 0 to 50, on the same datasets.
We observe that both hyper-parameters perform better on the \textit{SZ} dataset than on the \textit{MCM} dataset. 
The model is particularly sensitive to changes in $\delta$ within the range of 0 to 0.6, with the highest HR value occurring at $\delta$=0.6, indicating that both the \textit{MCM} and \textit{SZ} markets reach their optimal performance at this specific $\delta$ value. As for $\tau$, which controls the weight of $\mathcal{P}$ within the model, an optimal balance is achieved at $\tau=3$. A $\tau$ value set too high or too low disrupts this balance, causing the perspective matrix to completely dominate or be ignored by the model, making the model more subjective or too objective and ultimately leading to performance degradation.

\section{CONCLUSIONS}
Our research marks a significant advancement in the field of time series supplier allocation. By integrating the BL model with the innovative perspective matrix, we have successfully transformed this complex issue into a manageable risk optimization problem, capturing both temporal and spatial information in market dynamics. The implementation of the spatio-temporal graph neural network of our supplier relationship has notably streamlined the process, offering a more profound understanding of supplier inter-dependencies. The introduction of the masked ranking mechanism, specifically designed to address the lack of supervisory signals and unreliable issues, has further refined our approach, leading to more efficient supplier ranking and decision-making processes. Tested against two datasets, our \M~ not only demonstrates superior performance over the traditional models but also sets a new benchmark in the industry. Furthermore, this research paves the way for future developments in supply chain management, encouraging continued innovation and exploration in this vital field.


\balance
\bibliographystyle{ACM-Reference-Format}
\bibliography{acmart}


\begin{thebibliography}{70}


\ifx \showCODEN    \undefined \def \showCODEN     #1{\unskip}     \fi
\ifx \showDOI      \undefined \def \showDOI       #1{#1}\fi
\ifx \showISBNx    \undefined \def \showISBNx     #1{\unskip}     \fi
\ifx \showISBNxiii \undefined \def \showISBNxiii  #1{\unskip}     \fi
\ifx \showISSN     \undefined \def \showISSN      #1{\unskip}     \fi
\ifx \showLCCN     \undefined \def \showLCCN      #1{\unskip}     \fi
\ifx \shownote     \undefined \def \shownote      #1{#1}          \fi
\ifx \showarticletitle \undefined \def \showarticletitle #1{#1}   \fi
\ifx \showURL      \undefined \def \showURL       {\relax}        \fi
\providecommand\bibfield[2]{#2}
\providecommand\bibinfo[2]{#2}
\providecommand\natexlab[1]{#1}
\providecommand\showeprint[2][]{arXiv:#2}

\bibitem[Alikhani et~al\mbox{.}(2019)]%
        {Alikhani_Torabi_Altay_2019}
\bibfield{author}{\bibinfo{person}{Reza Alikhani}, \bibinfo{person}{S.Ali Torabi}, {and} \bibinfo{person}{Nezih Altay}.} \bibinfo{year}{2019}\natexlab{}.
\newblock \showarticletitle{Strategic supplier selection under sustainability and risk criteria}.
\newblock \bibinfo{journal}{\emph{Int J Prod Econ}} (\bibinfo{year}{2019}).
\newblock


\bibitem[Bayazit(2006)]%
        {Bayazit_2006}
\bibfield{author}{\bibinfo{person}{Ozden Bayazit}.} \bibinfo{year}{2006}\natexlab{}.
\newblock \showarticletitle{Use of analytic network process in vendor selection decisions}.
\newblock \bibinfo{journal}{\emph{BIJ}} (\bibinfo{year}{2006}).
\newblock


\bibitem[Bertsimas et~al\mbox{.}(2012)]%
        {bertsimas2012inverse}
\bibfield{author}{\bibinfo{person}{Dimitris Bertsimas}, \bibinfo{person}{Vishal Gupta}, {and} \bibinfo{person}{Ioannis~Ch Paschalidis}.} \bibinfo{year}{2012}\natexlab{}.
\newblock \showarticletitle{Inverse optimization: A new perspective on the Black-Litterman model}.
\newblock \bibinfo{journal}{\emph{Oper. Res}} (\bibinfo{year}{2012}), \bibinfo{pages}{1389--1403}.
\newblock


\bibitem[Black and Litterman(1991)]%
        {Black_Litterman_1991}
\bibfield{author}{\bibinfo{person}{Fischer Black} {and} \bibinfo{person}{Robert~B Litterman}.} \bibinfo{year}{1991}\natexlab{}.
\newblock \showarticletitle{Asset Allocation}.
\newblock \bibinfo{journal}{\emph{J. Fixed Income}} (\bibinfo{year}{1991}).
\newblock


\bibitem[Blondel et~al\mbox{.}(2021)]%
        {blondel2020fast}
\bibfield{author}{\bibinfo{person}{Mathieu Blondel}, \bibinfo{person}{Olivier Teboul}, \bibinfo{person}{Quentin Berthet}, {and} \bibinfo{person}{Josip Djolonga}.} \bibinfo{year}{2021}\natexlab{}.
\newblock \showarticletitle{Fast Differentiable Sorting and Ranking}. In \bibinfo{booktitle}{\emph{ICML}}.
\newblock


\bibitem[Bowser-Chao and Dzialo(1993)]%
        {Bowser_Chao_1993}
\bibfield{author}{\bibinfo{person}{David Bowser-Chao} {and} \bibinfo{person}{Debra~L. Dzialo}.} \bibinfo{year}{1993}\natexlab{}.
\newblock \showarticletitle{Comparison of the use of binary decision trees and neural networks in top-quark detection}.
\newblock \bibinfo{journal}{\emph{Phys. Rev. D}} (\bibinfo{year}{1993}).
\newblock


\bibitem[Brockmann et~al\mbox{.}(2022)]%
        {brockmann2022supply}
\bibfield{author}{\bibinfo{person}{Nils Brockmann}, \bibinfo{person}{Edward Elson~Kosasih}, {and} \bibinfo{person}{Alexandra Brintrup}.} \bibinfo{year}{2022}\natexlab{}.
\newblock \showarticletitle{Supply Chain Link Prediction on Uncertain Knowledge Graph}. In \bibinfo{booktitle}{\emph{SIGKDD}}.
\newblock


\bibitem[Chai and Ngai(2020)]%
        {Chai_Ngai_2020}
\bibfield{author}{\bibinfo{person}{Junyi Chai} {and} \bibinfo{person}{Eric~W.T. Ngai}.} \bibinfo{year}{2020}\natexlab{}.
\newblock \showarticletitle{Decision-making techniques in supplier selection: Recent accomplishments and what lies ahead}.
\newblock \bibinfo{journal}{\emph{Expert Syst. Appl}} (\bibinfo{year}{2020}).
\newblock


\bibitem[Chaising and Temdee(2017)]%
        {chaising2017application}
\bibfield{author}{\bibinfo{person}{Supansa Chaising} {and} \bibinfo{person}{Punnarumol Temdee}.} \bibinfo{year}{2017}\natexlab{}.
\newblock \showarticletitle{Application of a hybrid multi-criteria decision making approach for selecting of raw material supplier for small and medium enterprises}. In \bibinfo{booktitle}{\emph{ICDAMT}}.
\newblock


\bibitem[Chen et~al\mbox{.}(2006)]%
        {chen2006fuzzy}
\bibfield{author}{\bibinfo{person}{Chen-Tung Chen}, \bibinfo{person}{Ching-Torng Lin}, {and} \bibinfo{person}{Sue-Fn Huang}.} \bibinfo{year}{2006}\natexlab{}.
\newblock \showarticletitle{A fuzzy approach for supplier evaluation and selection in supply chain management}.
\newblock \bibinfo{journal}{\emph{Int J Prod Econ}} (\bibinfo{year}{2006}).
\newblock


\bibitem[Chen et~al\mbox{.}(2023)]%
        {mlpt1}
\bibfield{author}{\bibinfo{person}{Si-An Chen}, \bibinfo{person}{Chun-Liang Li}, \bibinfo{person}{Nate Yoder}, \bibinfo{person}{Sercan~O. Arik}, {and} \bibinfo{person}{Tomas Pfister}.} \bibinfo{year}{2023}\natexlab{}.
\newblock \bibinfo{title}{TSMixer: An All-MLP Architecture for Time Series Forecasting}.
\newblock
\newblock
\showeprint[arxiv]{2303.06053}~[cs.LG]


\bibitem[Chen et~al\mbox{.}(2020)]%
        {Chen_Xu_Han_Fu_Pi_Joe-Wong_Li_Zhang_Noh_Zhang_2020}
\bibfield{author}{\bibinfo{person}{Xinlei Chen}, \bibinfo{person}{Susu Xu}, \bibinfo{person}{Jun Han}, \bibinfo{person}{Haohao Fu}, \bibinfo{person}{Xidong Pi}, \bibinfo{person}{Carlee Joe-Wong}, \bibinfo{person}{Yong Li}, \bibinfo{person}{Lin Zhang}, \bibinfo{person}{HaeYoung Noh}, {and} \bibinfo{person}{Pei Zhang}.} \bibinfo{year}{2020}\natexlab{}.
\newblock \showarticletitle{PAS: Prediction-Based Actuation System for City-Scale Ridesharing Vehicular Mobile Crowdsensing}.
\newblock \bibinfo{journal}{\emph{IEEE Internet Things J}} (\bibinfo{year}{2020}).
\newblock


\bibitem[Choy et~al\mbox{.}(2005)]%
        {Choy_Lee_Lau_Choy_2005}
\bibfield{author}{\bibinfo{person}{K.L. Choy}, \bibinfo{person}{W.B. Lee}, \bibinfo{person}{Henry~C.W. Lau}, {and} \bibinfo{person}{L.C. Choy}.} \bibinfo{year}{2005}\natexlab{}.
\newblock \showarticletitle{A knowledge-based supplier intelligence retrieval system for outsource manufacturing}.
\newblock \bibinfo{journal}{\emph{Knowl Based Syst}} (\bibinfo{year}{2005}).
\newblock


\bibitem[De~Boer et~al\mbox{.}(2005)]%
        {de2005tutorial}
\bibfield{author}{\bibinfo{person}{Pieter-Tjerk De~Boer}, \bibinfo{person}{Dirk~P Kroese}, \bibinfo{person}{Shie Mannor}, {and} \bibinfo{person}{Reuven~Y Rubinstein}.} \bibinfo{year}{2005}\natexlab{}.
\newblock \showarticletitle{A tutorial on the cross-entropy method}.
\newblock \bibinfo{journal}{\emph{ANN OPER RES}} (\bibinfo{year}{2005}).
\newblock


\bibitem[Dotoli et~al\mbox{.}(2015)]%
        {dotoli2015integrated}
\bibfield{author}{\bibinfo{person}{Mariagrazia Dotoli}, \bibinfo{person}{Nicola Epicoco}, {and} \bibinfo{person}{Marco Falagario}.} \bibinfo{year}{2015}\natexlab{}.
\newblock \showarticletitle{Integrated supplier selection and order allocation under uncertainty in agile supply chains}. In \bibinfo{booktitle}{\emph{ETFA}}.
\newblock


\bibitem[Duică et~al\mbox{.}(2018)]%
        {Duică2018Selecting}
\bibfield{author}{\bibinfo{person}{M. Duică}, \bibinfo{person}{N. Florea}, {and} \bibinfo{person}{Anișoara Duică}.} \bibinfo{year}{2018}\natexlab{}.
\newblock \showarticletitle{Selecting the Right Suppliers in Procurement Process along Supply Chain-a Mathematical Modeling Approach}.
\newblock \bibinfo{journal}{\emph{Valahian J. Econom. Stud.}} (\bibinfo{year}{2018}).
\newblock


\bibitem[Fabozzi et~al\mbox{.}(2015)]%
        {Fabozzi_Markowitz_Gupta_2015}
\bibfield{author}{\bibinfo{person}{Frank~J. Fabozzi}, \bibinfo{person}{Harry~M. Markowitz}, {and} \bibinfo{person}{Francis Gupta}.} \bibinfo{year}{2015}\natexlab{}.
\newblock \bibinfo{title}{Portfolio Selection}.
\newblock
\newblock


\bibitem[Fang et~al\mbox{.}(2023)]%
        {fang2023spatio}
\bibfield{author}{\bibinfo{person}{Yuchen Fang}, \bibinfo{person}{Yanjun Qin}, \bibinfo{person}{Haiyong Luo}, \bibinfo{person}{Fang Zhao}, \bibinfo{person}{Bingbing Xu}, \bibinfo{person}{Liang Zeng}, {and} \bibinfo{person}{Chenxing Wang}.} \bibinfo{year}{2023}\natexlab{}.
\newblock \showarticletitle{When Spatio-Temporal Meet Wavelets: Disentangled Traffic Forecasting via Efficient Spectral Graph Attention Networks}. In \bibinfo{booktitle}{\emph{ICDE}}.
\newblock


\bibitem[Guo et~al\mbox{.}(2019)]%
        {10.1609/aaai.v33i01.3301922}
\bibfield{author}{\bibinfo{person}{Shengnan Guo}, \bibinfo{person}{Youfang Lin}, \bibinfo{person}{Ning Feng}, \bibinfo{person}{Chao Song}, {and} \bibinfo{person}{Huaiyu Wan}.} \bibinfo{year}{2019}\natexlab{}.
\newblock \showarticletitle{Attention based spatial-temporal graph convolutional networks for traffic flow forecasting}. In \bibinfo{booktitle}{\emph{AAAI}}.
\newblock


\bibitem[Gören(2018)]%
        {Gören_2018}
\bibfield{author}{\bibinfo{person}{Hacer~Güner Gören}.} \bibinfo{year}{2018}\natexlab{}.
\newblock \showarticletitle{A decision framework for sustainable supplier selection and order allocation with lost sales}.
\newblock \bibinfo{journal}{\emph{J. Clean}} (\bibinfo{year}{2018}).
\newblock


\bibitem[Hamdan and Jarndal(2017)]%
        {hamdan2017two}
\bibfield{author}{\bibinfo{person}{Sadeque Hamdan} {and} \bibinfo{person}{Anwar Jarndal}.} \bibinfo{year}{2017}\natexlab{}.
\newblock \showarticletitle{A two stage green supplier selection and order allocation using AHP and multi-objective genetic algorithm optimization}. In \bibinfo{booktitle}{\emph{ICMSAO}}.
\newblock


\bibitem[Hao et~al\mbox{.}(2022)]%
        {hao2022three}
\bibfield{author}{\bibinfo{person}{Shiqi Hao}, \bibinfo{person}{Yang Liu}, \bibinfo{person}{Yu Wang}, \bibinfo{person}{Yuan Wang}, {and} \bibinfo{person}{Wenming Zhe}.} \bibinfo{year}{2022}\natexlab{}.
\newblock \showarticletitle{Three-stage root cause analysis for logistics time efficiency via explainable machine learning}. In \bibinfo{booktitle}{\emph{SIGKDD}}.
\newblock


\bibitem[Hasan et~al\mbox{.}(2020)]%
        {hasan2020resilient}
\bibfield{author}{\bibinfo{person}{Md~Mahmudul Hasan}, \bibinfo{person}{Dizuo Jiang}, \bibinfo{person}{AMM~Sharif Ullah}, {and} \bibinfo{person}{Md Noor-E-Alam}.} \bibinfo{year}{2020}\natexlab{}.
\newblock \showarticletitle{Resilient supplier selection in logistics 4.0 with heterogeneous information}.
\newblock \bibinfo{journal}{\emph{EXPERT SYST APPL}} (\bibinfo{year}{2020}).
\newblock


\bibitem[Ho et~al\mbox{.}(2010)]%
        {Ho_Xu_Dey_2010}
\bibfield{author}{\bibinfo{person}{William Ho}, \bibinfo{person}{Xiaowei Xu}, {and} \bibinfo{person}{Prasanta~K. Dey}.} \bibinfo{year}{2010}\natexlab{}.
\newblock \showarticletitle{Multi-criteria decision making approaches for supplier evaluation and selection: A literature review}.
\newblock \bibinfo{journal}{\emph{Eur. J. Oper. Res}} (\bibinfo{year}{2010}).
\newblock


\bibitem[Hui et~al\mbox{.}(2023a)]%
        {Hui_Fang_Xia_Aykent_Ku_2023}
\bibfield{author}{\bibinfo{person}{Bo Hui}, \bibinfo{person}{Yuchen Fang}, \bibinfo{person}{Tian Xia}, \bibinfo{person}{Sarp Aykent}, {and} \bibinfo{person}{Wei-Shinn Ku}.} \bibinfo{year}{2023}\natexlab{a}.
\newblock \showarticletitle{Constrained Market Share Maximization by Signal-Guided Optimization}. In \bibinfo{booktitle}{\emph{AAAI}}.
\newblock


\bibitem[Hui et~al\mbox{.}(2023b)]%
        {hui2023constrained}
\bibfield{author}{\bibinfo{person}{Bo Hui}, \bibinfo{person}{Yuchen Fang}, \bibinfo{person}{Tian Xia}, \bibinfo{person}{Sarp Aykent}, {and} \bibinfo{person}{Wei-Shinn Ku}.} \bibinfo{year}{2023}\natexlab{b}.
\newblock \showarticletitle{Constrained market share maximization by signal-guided optimization}. In \bibinfo{booktitle}{\emph{AAAI}}.
\newblock


\bibitem[Islam et~al\mbox{.}(2021)]%
        {islam2021machine}
\bibfield{author}{\bibinfo{person}{Samiul Islam}, \bibinfo{person}{Saman~Hassanzadeh Amin}, {and} \bibinfo{person}{Leslie~J Wardley}.} \bibinfo{year}{2021}\natexlab{}.
\newblock \showarticletitle{Machine learning and optimization models for supplier selection and order allocation planning}.
\newblock \bibinfo{journal}{\emph{Int J Prod Econ}} (\bibinfo{year}{2021}).
\newblock


\bibitem[Jayaraman et~al\mbox{.}(1999)]%
        {jayaraman1999supplier}
\bibfield{author}{\bibinfo{person}{Vaidyanathan Jayaraman}, \bibinfo{person}{Rajesh Srivastava}, {and} \bibinfo{person}{WC Benton}.} \bibinfo{year}{1999}\natexlab{}.
\newblock \showarticletitle{Supplier selection and order quantity allocation: a comprehensive model}.
\newblock \bibinfo{journal}{\emph{J. Supply Chain Manag}} (\bibinfo{year}{1999}).
\newblock


\bibitem[Jiang et~al\mbox{.}(2024)]%
        {katz}
\bibfield{author}{\bibinfo{person}{Xinke Jiang}, \bibinfo{person}{Zidi Qin}, \bibinfo{person}{Jiarong Xu}, {and} \bibinfo{person}{Xiang Ao}.} \bibinfo{year}{2024}\natexlab{}.
\newblock \showarticletitle{Incomplete Graph Learning via Attribute-Structure Decoupled Variational Auto-Encoder}. In \bibinfo{booktitle}{\emph{WSDM}}.
\newblock


\bibitem[Jiang et~al\mbox{.}(2023)]%
        {10.1145/3583780.3615215}
\bibfield{author}{\bibinfo{person}{Xinke Jiang}, \bibinfo{person}{Dingyi Zhuang}, \bibinfo{person}{Xianghui Zhang}, \bibinfo{person}{Hao Chen}, \bibinfo{person}{Jiayuan Luo}, {and} \bibinfo{person}{Xiaowei Gao}.} \bibinfo{year}{2023}\natexlab{}.
\newblock \showarticletitle{Uncertainty Quantification via Spatial-Temporal Tweedie Model for Zero-inflated and Long-tail Travel Demand Prediction}. In \bibinfo{booktitle}{\emph{CIKM}}.
\newblock


\bibitem[Kara et~al\mbox{.}(2019)]%
        {kara2019hybrid}
\bibfield{author}{\bibinfo{person}{Mahmut Kara}, \bibinfo{person}{Aydin Ulucan}, {and} \bibinfo{person}{Kazim~Baris Atici}.} \bibinfo{year}{2019}\natexlab{}.
\newblock \showarticletitle{A hybrid approach for generating investor views in Black--Litterman model}.
\newblock \bibinfo{journal}{\emph{Expert Syst. Appl.}} (\bibinfo{year}{2019}).
\newblock


\bibitem[Kawtummachai and Van~Hop(2005)]%
        {kawtummachai2005order}
\bibfield{author}{\bibinfo{person}{Ruengsak Kawtummachai} {and} \bibinfo{person}{Nguyen Van~Hop}.} \bibinfo{year}{2005}\natexlab{}.
\newblock \showarticletitle{Order allocation in a multiple-supplier environment}.
\newblock \bibinfo{journal}{\emph{Int J Prod Econ}} (\bibinfo{year}{2005}).
\newblock


\bibitem[Kolm et~al\mbox{.}(2021)]%
        {Kolm_Ritter_Simonian_2021}
\bibfield{author}{\bibinfo{person}{PetterN. Kolm}, \bibinfo{person}{Gordon Ritter}, {and} \bibinfo{person}{Joseph Simonian}.} \bibinfo{year}{2021}\natexlab{}.
\newblock \showarticletitle{Black-Litterman and Beyond: The Bayesian Paradigm in Investment Management}.
\newblock \bibinfo{journal}{\emph{SSRN}} (\bibinfo{year}{2021}).
\newblock


\bibitem[Kuroiwa and Beck(2023)]%
        {Kuroiwa_2023}
\bibfield{author}{\bibinfo{person}{Ryo Kuroiwa} {and} \bibinfo{person}{J.~Christopher Beck}.} \bibinfo{year}{2023}\natexlab{}.
\newblock \showarticletitle{Domain-Independent Dynamic Programming: Generic State Space Search for Combinatorial Optimization}. In \bibinfo{booktitle}{\emph{ICAPS}}.
\newblock


\bibitem[Lea et~al\mbox{.}(2017)]%
        {lea2017temporal}
\bibfield{author}{\bibinfo{person}{Colin Lea}, \bibinfo{person}{Michael~D Flynn}, \bibinfo{person}{Rene Vidal}, \bibinfo{person}{Austin Reiter}, {and} \bibinfo{person}{Gregory~D Hager}.} \bibinfo{year}{2017}\natexlab{}.
\newblock \showarticletitle{Temporal convolutional networks for action segmentation and detection}. In \bibinfo{booktitle}{\emph{CVPR}}.
\newblock


\bibitem[LeCun et~al\mbox{.}(2015)]%
        {lecun2015deep}
\bibfield{author}{\bibinfo{person}{Yann LeCun}, \bibinfo{person}{Yoshua Bengio}, {and} \bibinfo{person}{Geoffrey Hinton}.} \bibinfo{year}{2015}\natexlab{}.
\newblock \showarticletitle{Deep learning}.
\newblock \bibinfo{journal}{\emph{nature}} (\bibinfo{year}{2015}).
\newblock


\bibitem[Lee(2009)]%
        {lee2009fuzzy}
\bibfield{author}{\bibinfo{person}{Amy~HI Lee}.} \bibinfo{year}{2009}\natexlab{}.
\newblock \showarticletitle{A fuzzy supplier selection model with the consideration of benefits, opportunities, costs and risks}.
\newblock \bibinfo{journal}{\emph{Expert Syst. Appl}} (\bibinfo{year}{2009}).
\newblock


\bibitem[Li et~al\mbox{.}(2021)]%
        {li2021large}
\bibfield{author}{\bibinfo{person}{Duanshun Li}, \bibinfo{person}{Jing Liu}, \bibinfo{person}{Jinsung Jeon}, \bibinfo{person}{Seoyoung Hong}, \bibinfo{person}{Thai Le}, \bibinfo{person}{Dongwon Lee}, {and} \bibinfo{person}{Noseong Park}.} \bibinfo{year}{2021}\natexlab{}.
\newblock \showarticletitle{Large-Scale Data-Driven Airline Market Influence Maximization}. In \bibinfo{booktitle}{\emph{SIGKDD}}.
\newblock


\bibitem[Li et~al\mbox{.}(2022)]%
        {SPGCL}
\bibfield{author}{\bibinfo{person}{Rongfan Li}, \bibinfo{person}{Ting Zhong}, \bibinfo{person}{Xinke Jiang}, \bibinfo{person}{Goce Trajcevski}, \bibinfo{person}{Jin Wu}, {and} \bibinfo{person}{Fan Zhou}.} \bibinfo{year}{2022}\natexlab{}.
\newblock \showarticletitle{Mining Spatio-Temporal Relations via Self-Paced Graph Contrastive Learning}. In \bibinfo{booktitle}{\emph{SIGKDD}}.
\newblock


\bibitem[Liao and Rittscher(2007)]%
        {Liao_Rittscher_2007}
\bibfield{author}{\bibinfo{person}{Zhiying Liao} {and} \bibinfo{person}{Jens Rittscher}.} \bibinfo{year}{2007}\natexlab{}.
\newblock \showarticletitle{A multi-objective supplier selection model under stochastic demand conditions}.
\newblock \bibinfo{journal}{\emph{Int J Prod Econ}} (\bibinfo{year}{2007}).
\newblock


\bibitem[Lyu et~al\mbox{.}(2023)]%
        {Lyu_Wang_Song_Liu_He_Zhang_2023}
\bibfield{author}{\bibinfo{person}{Wenjun Lyu}, \bibinfo{person}{Haotian Wang}, \bibinfo{person}{Yiwei Song}, \bibinfo{person}{Yunhuai Liu}, \bibinfo{person}{Tian He}, {and} \bibinfo{person}{Desheng Zhang}.} \bibinfo{year}{2023}\natexlab{}.
\newblock \showarticletitle{A Prediction-and-Scheduling Framework for Efficient Order Transfer in Logistics}. In \bibinfo{booktitle}{\emph{IJCAI}}.
\newblock


\bibitem[Markowitz(1952)]%
        {capm1}
\bibfield{author}{\bibinfo{person}{Harry Markowitz}.} \bibinfo{year}{1952}\natexlab{}.
\newblock \showarticletitle{Portfolio Selection}.
\newblock \bibinfo{journal}{\emph{J Finance}} (\bibinfo{year}{1952}).
\newblock


\bibitem[Mitra and Zhang(2014)]%
        {mitra2014multivariate}
\bibfield{author}{\bibinfo{person}{Ritwik Mitra} {and} \bibinfo{person}{Cun-Hui Zhang}.} \bibinfo{year}{2014}\natexlab{}.
\newblock \bibinfo{title}{Multivariate Analysis of Nonparametric Estimates of Large Correlation Matrices}.
\newblock
\newblock
\showeprint[arxiv]{1403.6195}~[math.ST]


\bibitem[Moheb-Alizadeh and Handfield(2019)]%
        {moheb2019sustainable}
\bibfield{author}{\bibinfo{person}{Hadi Moheb-Alizadeh} {and} \bibinfo{person}{Robert Handfield}.} \bibinfo{year}{2019}\natexlab{}.
\newblock \showarticletitle{Sustainable supplier selection and order allocation: A novel multi-objective programming model with a hybrid solution approach}.
\newblock \bibinfo{journal}{\emph{Comput Ind Eng}} (\bibinfo{year}{2019}).
\newblock


\bibitem[Mossin(1966)]%
        {capm3}
\bibfield{author}{\bibinfo{person}{Jan Mossin}.} \bibinfo{year}{1966}\natexlab{}.
\newblock \showarticletitle{EQUILIBRIUM IN A CAPITAL ASSET MARKET}.
\newblock \bibinfo{journal}{\emph{Econometrica}} (\bibinfo{year}{1966}).
\newblock


\bibitem[Nasiri et~al\mbox{.}(2018)]%
        {nasiri2018incorporating}
\bibfield{author}{\bibinfo{person}{Mohammad~Mahdi Nasiri}, \bibinfo{person}{Ali Rahbari}, \bibinfo{person}{Frank Werner}, {and} \bibinfo{person}{Roya Karimi}.} \bibinfo{year}{2018}\natexlab{}.
\newblock \showarticletitle{Incorporating supplier selection and order allocation into the vehicle routing and multi-cross-dock scheduling problem}.
\newblock \bibinfo{journal}{\emph{Int. J. Prod. Res}} (\bibinfo{year}{2018}).
\newblock


\bibitem[Noori-Daryan et~al\mbox{.}(2019)]%
        {noori2019analyzing}
\bibfield{author}{\bibinfo{person}{Mahsa Noori-Daryan}, \bibinfo{person}{Ata~Allah Taleizadeh}, {and} \bibinfo{person}{Fariborz Jolai}.} \bibinfo{year}{2019}\natexlab{}.
\newblock \showarticletitle{Analyzing pricing, promised delivery lead time, supplier-selection, and ordering decisions of a multi-national supply chain under uncertain environment}.
\newblock \bibinfo{journal}{\emph{Int J Prod Econ}} (\bibinfo{year}{2019}).
\newblock


\bibitem[Paszke et~al\mbox{.}(2019)]%
        {paszke2019pytorch}
\bibfield{author}{\bibinfo{person}{Adam Paszke}, \bibinfo{person}{Sam Gross}, \bibinfo{person}{Francisco Massa}, \bibinfo{person}{Adam Lerer}, \bibinfo{person}{James Bradbury}, \bibinfo{person}{Gregory Chanan}, \bibinfo{person}{Trevor Killeen}, \bibinfo{person}{Zeming Lin}, \bibinfo{person}{Natalia Gimelshein}, \bibinfo{person}{Luca Antiga}, {et~al\mbox{.}}} \bibinfo{year}{2019}\natexlab{}.
\newblock \showarticletitle{Pytorch: An imperative style, high-performance deep learning library}. In \bibinfo{booktitle}{\emph{NeurIPS}}.
\newblock


\bibitem[Puvska and Stojanovic(2022)]%
        {Puka2022FuzzyMA}
\bibfield{author}{\bibinfo{person}{Adis Puvska} {and} \bibinfo{person}{Ilija Stojanovic}.} \bibinfo{year}{2022}\natexlab{}.
\newblock \showarticletitle{Fuzzy Multi-Criteria Analyses on Green Supplier Selection in an Agri-Food Company}.
\newblock \bibinfo{journal}{\emph{JIMD}} (\bibinfo{year}{2022}).
\newblock


\bibitem[Ren et~al\mbox{.}(2022)]%
        {ren2022better}
\bibfield{author}{\bibinfo{person}{Yi Ren}, \bibinfo{person}{Shangmin Guo}, {and} \bibinfo{person}{Danica~J Sutherland}.} \bibinfo{year}{2022}\natexlab{}.
\newblock \showarticletitle{Better supervisory signals by observing learning paths}.
\newblock \bibinfo{journal}{\emph{arXiv preprint arXiv:2203.02485}} (\bibinfo{year}{2022}).
\newblock


\bibitem[Rezaei et~al\mbox{.}(2020)]%
        {rezaei2020supplier}
\bibfield{author}{\bibinfo{person}{Aida Rezaei}, \bibinfo{person}{Masoud Rahiminezhad~Galankashi}, \bibinfo{person}{Shiva Mansoorzadeh}, {and} \bibinfo{person}{Farimah Mokhatab~Rafiei}.} \bibinfo{year}{2020}\natexlab{}.
\newblock \showarticletitle{Supplier selection and order allocation with lean manufacturing criteria: an integrated MCDM and Bi-objective modelling approach}.
\newblock \bibinfo{journal}{\emph{ENG MANAG J}} (\bibinfo{year}{2020}).
\newblock


\bibitem[Rosenblatt(1963)]%
        {MLP}
\bibfield{author}{\bibinfo{person}{Frank Rosenblatt}.} \bibinfo{year}{1963}\natexlab{}.
\newblock \showarticletitle{PRINCIPLES OF NEURODYNAMICS. PERCEPTRONS AND THE THEORY OF BRAIN MECHANISMS}.
\newblock \bibinfo{journal}{\emph{AJP}} (\bibinfo{year}{1963}).
\newblock


\bibitem[Sarkar and Mohapatra(2006)]%
        {Sarkar_Mohapatra_2006}
\bibfield{author}{\bibinfo{person}{Ashutosh Sarkar} {and} \bibinfo{person}{Pratap~K.J. Mohapatra}.} \bibinfo{year}{2006}\natexlab{}.
\newblock \showarticletitle{Evaluation of supplier capability and performance: A method for supply base reduction}.
\newblock \bibinfo{journal}{\emph{J. Purch}} (\bibinfo{year}{2006}).
\newblock


\bibitem[Sharpe(1963)]%
        {Sharpe1963A}
\bibfield{author}{\bibinfo{person}{W. Sharpe}.} \bibinfo{year}{1963}\natexlab{}.
\newblock \showarticletitle{A Simplified Model for Portfolio Analysis}.
\newblock \bibinfo{journal}{\emph{Manage Sci}} (\bibinfo{year}{1963}).
\newblock


\bibitem[Steinbach(2001)]%
        {Steinbach_2001}
\bibfield{author}{\bibinfo{person}{Marc~C. Steinbach}.} \bibinfo{year}{2001}\natexlab{}.
\newblock \showarticletitle{Markowitz Revisited: Mean-Variance Models in Financial Portfolio Analysis}.
\newblock \bibinfo{journal}{\emph{SIAM REV}} (\bibinfo{year}{2001}).
\newblock


\bibitem[Tan et~al\mbox{.}(1998)]%
        {tan1998supply}
\bibfield{author}{\bibinfo{person}{Keah-Choon Tan}, \bibinfo{person}{Vijay~R Kannan}, {and} \bibinfo{person}{Robert~B Handfield}.} \bibinfo{year}{1998}\natexlab{}.
\newblock \showarticletitle{Supply chain management: supplier performance and firm performance}.
\newblock \bibinfo{journal}{\emph{Int. J. Purch. Mater. Manage.}} (\bibinfo{year}{1998}).
\newblock


\bibitem[Tibshirani(1996)]%
        {Tibshirani1996}
\bibfield{author}{\bibinfo{person}{Robert Tibshirani}.} \bibinfo{year}{1996}\natexlab{}.
\newblock \showarticletitle{Regression Shrinkage and Selection Via the Lasso}.
\newblock \bibinfo{journal}{\emph{J R STAT SOC B}} (\bibinfo{year}{1996}).
\newblock


\bibitem[Treynor(1962)]%
        {capm2}
\bibfield{author}{\bibinfo{person}{Jack~L. Treynor}.} \bibinfo{year}{1962}\natexlab{}.
\newblock \bibinfo{title}{Toward a Theory of Market Value of Risky Assets}.
\newblock
\newblock


\bibitem[Veličković et~al\mbox{.}(2018)]%
        {GAT}
\bibfield{author}{\bibinfo{person}{Petar Veličković}, \bibinfo{person}{Guillem Cucurull}, \bibinfo{person}{Arantxa Casanova}, \bibinfo{person}{Adriana Romero}, \bibinfo{person}{Pietro Liò}, {and} \bibinfo{person}{Yoshua Bengio}.} \bibinfo{year}{2018}\natexlab{}.
\newblock \showarticletitle{Graph Attention Networks}. In \bibinfo{booktitle}{\emph{ICLR}}.
\newblock


\bibitem[Wan et~al\mbox{.}(2023)]%
        {Wan_Rao_Dong_2023}
\bibfield{author}{\bibinfo{person}{Shu-Ping Wan}, \bibinfo{person}{Tian Rao}, {and} \bibinfo{person}{Jiu-Ying Dong}.} \bibinfo{year}{2023}\natexlab{}.
\newblock \showarticletitle{Time-series based multi-criteria large-scale group decision making with intuitionistic fuzzy information and application to multi-period battery supplier selection}.
\newblock \bibinfo{journal}{\emph{Expert Syst. Appl}} (\bibinfo{year}{2023}).
\newblock


\bibitem[Wang and Bovik(2009)]%
        {wang2009mean}
\bibfield{author}{\bibinfo{person}{Zhou Wang} {and} \bibinfo{person}{Alan~C Bovik}.} \bibinfo{year}{2009}\natexlab{}.
\newblock \showarticletitle{Mean squared error: Love it or leave it? A new look at signal fidelity measures}.
\newblock \bibinfo{journal}{\emph{IEEE Signal Process Mag}} (\bibinfo{year}{2009}).
\newblock


\bibitem[Way et~al\mbox{.}(2018)]%
        {way2018wright}
\bibfield{author}{\bibinfo{person}{Rupert Way}, \bibinfo{person}{François Lafond}, \bibinfo{person}{Fabrizio Lillo}, \bibinfo{person}{Valentyn Panchenko}, {and} \bibinfo{person}{J.~Doyne Farmer}.} \bibinfo{year}{2018}\natexlab{}.
\newblock \bibinfo{title}{Wright meets Markowitz: How standard portfolio theory changes when assets are technologies following experience curves}.
\newblock
\newblock
\showeprint[arxiv]{1705.03423}~[q-fin.EC]


\bibitem[Wen et~al\mbox{.}(2022)]%
        {Wen_Lin_Mao_Wu_Zhao_Wang_Zheng_Wu_Hu_Wan_2022}
\bibfield{author}{\bibinfo{person}{Haomin Wen}, \bibinfo{person}{Youfang Lin}, \bibinfo{person}{Xiaowei Mao}, \bibinfo{person}{Fan Wu}, \bibinfo{person}{Yiji Zhao}, \bibinfo{person}{Haochen Wang}, \bibinfo{person}{Jianbin Zheng}, \bibinfo{person}{Lixia Wu}, \bibinfo{person}{Haoyuan Hu}, {and} \bibinfo{person}{Huaiyu Wan}.} \bibinfo{year}{2022}\natexlab{}.
\newblock \showarticletitle{Graph2Route: A Dynamic Spatial-Temporal Graph Neural Network for Pick-up and Delivery Route Prediction}. In \bibinfo{booktitle}{\emph{SIGKDD}}.
\newblock


\bibitem[Wu et~al\mbox{.}(2021a)]%
        {wu2021inductive}
\bibfield{author}{\bibinfo{person}{Yuankai Wu}, \bibinfo{person}{Dingyi Zhuang}, \bibinfo{person}{Aurelie Labbe}, {and} \bibinfo{person}{Lijun Sun}.} \bibinfo{year}{2021}\natexlab{a}.
\newblock \showarticletitle{Inductive Graph Neural Networks for Spatiotemporal Kriging}. In \bibinfo{booktitle}{\emph{AAAI}}.
\newblock


\bibitem[Wu et~al\mbox{.}(2021b)]%
        {wu2021spatial}
\bibfield{author}{\bibinfo{person}{Yuankai Wu}, \bibinfo{person}{Dingyi Zhuang}, \bibinfo{person}{Mengying Lei}, \bibinfo{person}{Aurelie Labbe}, {and} \bibinfo{person}{Lijun Sun}.} \bibinfo{year}{2021}\natexlab{b}.
\newblock \showarticletitle{Spatial Aggregation and Temporal Convolution Networks for Real-time Kriging}.
\newblock \bibinfo{journal}{\emph{arXiv preprint arXiv:2109.12144}} (\bibinfo{year}{2021}).
\newblock


\bibitem[Xing et~al\mbox{.}(2018)]%
        {Xing_Cambria_Malandri_Vercellis_2018}
\bibfield{author}{\bibinfo{person}{FrankZ. Xing}, \bibinfo{person}{Erik Cambria}, \bibinfo{person}{Lorenzo Malandri}, {and} \bibinfo{person}{Carlo Vercellis}.} \bibinfo{year}{2018}\natexlab{}.
\newblock \showarticletitle{Discovering Bayesian Market Views for Intelligent Asset Allocation}.
\newblock \bibinfo{journal}{\emph{RePEc}} (\bibinfo{year}{2018}).
\newblock


\bibitem[Zhang et~al\mbox{.}(2024)]%
        {gpfn}
\bibfield{author}{\bibinfo{person}{Ruizhe Zhang}, \bibinfo{person}{Xinke Jiang}, \bibinfo{person}{Yuchen Fang}, \bibinfo{person}{Jiayuan Luo}, \bibinfo{person}{Yongxin Xu}, \bibinfo{person}{Yichen Zhu}, \bibinfo{person}{Xu Chu}, \bibinfo{person}{Junfeng Zhao}, {and} \bibinfo{person}{Yasha Wang}.} \bibinfo{year}{2024}\natexlab{}.
\newblock \bibinfo{title}{Infinite-Horizon Graph Filters: Leveraging Power Series to Enhance Sparse Information Aggregation}.
\newblock
\newblock
\showeprint[arxiv]{2401.09943}~[cs.LG]


\bibitem[Zhang et~al\mbox{.}(2022)]%
        {mlpt2}
\bibfield{author}{\bibinfo{person}{Tianping Zhang}, \bibinfo{person}{Yizhuo Zhang}, \bibinfo{person}{Wei Cao}, \bibinfo{person}{Jiang Bian}, \bibinfo{person}{Xiaohan Yi}, \bibinfo{person}{Shun Zheng}, {and} \bibinfo{person}{Jian Li}.} \bibinfo{year}{2022}\natexlab{}.
\newblock \bibinfo{title}{Less Is More: Fast Multivariate Time Series Forecasting with Light Sampling-oriented MLP Structures}.
\newblock
\newblock
\showeprint[arxiv]{2207.01186}~[cs.LG]


\bibitem[Zhao et~al\mbox{.}(2023)]%
        {9835149}
\bibfield{author}{\bibinfo{person}{Fuqing Zhao}, \bibinfo{person}{Zesong Xu}, \bibinfo{person}{Ling Wang}, \bibinfo{person}{Ningning Zhu}, \bibinfo{person}{Tianpeng Xu}, {and} \bibinfo{person}{J. Jonrinaldi}.} \bibinfo{year}{2023}\natexlab{}.
\newblock \showarticletitle{A Population-Based Iterated Greedy Algorithm for Distributed Assembly No-Wait Flow-Shop Scheduling Problem}.
\newblock \bibinfo{journal}{\emph{IEEE Trans Industr Inform}} (\bibinfo{year}{2023}).
\newblock


\bibitem[Zhuang et~al\mbox{.}(2022)]%
        {zhuang2022uncertainty}
\bibfield{author}{\bibinfo{person}{Dingyi Zhuang}, \bibinfo{person}{Shenhao Wang}, \bibinfo{person}{Haris Koutsopoulos}, {and} \bibinfo{person}{Jinhua Zhao}.} \bibinfo{year}{2022}\natexlab{}.
\newblock \showarticletitle{Uncertainty Quantification of Sparse Travel Demand Prediction with Spatial-Temporal Graph Neural Networks}. In \bibinfo{booktitle}{\emph{SIGKDD}}.
\newblock


\end{thebibliography}

\newpage
\balance
\appendix
\section{Optimization Transfer}
\label{AP: Optimization Transfer}
As previously outlined, the original TSSA optimization problem is presented as follows:
\begin{align}
    \begin{cases}
        \max  & \underbrace{\sum_{i=1}^{N} \mathcal{B}_{it} \mu_i}_{\text{Profit Item}} - \delta \underbrace{\sum_{i=1}^{N} \mathcal{B}_{it}^2 (\mathcal{O}_{it} - \mathcal{S}_{it})^\kappa}_{\text{Risk Item}}  \\ 
        \text{subject to} &\sum_{i=1}^{N} \mathcal{B}_{it} = M, \quad \forall i=1,2,\cdots,N, \quad \mathcal{B}_{it} \geq 0
    \end{cases}.
\end{align}
To align this with more conventional portfolio optimization models and facilitate comparisons, we define a weight coefficient \(w_{it}\) for supplier \(\mathcal{SU}_i\) at time slot \(t\) as \(w_{it}=\frac{\mathcal{B}_{it}}{M}\), where $M$ remains the total volume constant.

This introduction accounts for the influence of the total investment amount \(M\) on the optimization problem, allowing for a simplification that renders the objective function more straightforward. By substituting \(w_{it}=\frac{\mathcal{B}_{it}}{M}\) into the equation, the optimization can be rewritten as:
\begin{equation}
 \begin{aligned}
& \max \quad M \left( \sum_{i=1}^{N} w_{it} \mu_i - \delta M \sum_{i=1}^{N} w_{it}^2 (\mathcal{O}_{it} - \mathcal{S}_{it})^\kappa \right) \\
\sim \quad &
\max  \quad 
\sum_{i=1}^{N} w_{it} \mu_i - \frac{\delta}{2}\sum_{i=1}^{N} w_{it}^2 (\mathcal{O}_{it} - \mathcal{S}_{it})^\kappa  \\
= \quad &
\max  \quad 
\mathcal{W}_{t} \mu - \frac{\delta}{2} \mathcal{W}_{t}^T (\mathcal{O}_{t} - \mathcal{S}_{t})^\kappa \mathcal{W}_{t},
 \end{aligned}
\end{equation}
where $\mathcal{W}_{t}$ is the weight allocation matrix with entry $w_{it}=\mathcal{B}_{it}/M$ signifies the proportion of total material volume allocated to supplier $\mathcal{SU}_i$ at time $t$.

Thus, we rewrite the objective as:
\begin{align}
    \begin{cases}
        \max  & \underbrace{\vphantom{\mathcal{W}_{t}^T (\mathcal{O}_{t} - \mathcal{S}_{t})^\kappa \mathcal{W}_{t}}\mathcal{W}_{t} \mu}_{\text{Profit Item}} - \frac{\delta}{2} \underbrace{ \mathcal{W}_{t}^T (\mathcal{O}_{t} - \mathcal{S}_{t})^\kappa \mathcal{W}_{t}}_{\text{Risk Item}}  \\ 
        \text{subject to} &\sum_{i=1}^{N} w_{it} = 1, \quad \forall i=1,2,\cdots,N, \quad 1 \ge w_{it} \geq 0.
    \end{cases}.
\end{align}
The adjustment yields a more straightforward formula, enhancing our objective's clarity and comparability with standard models.

\section{Revisiting Black-Litterman Model}
\label{ap: revisit}
The Black-Litterman (BL) model, a sophisticated portfolio optimization framework introduced by Fischer Black and Robert Litterman in 1991, integrates market equilibrium expectations with individual investor views to derive an optimized asset allocation strategy. This model significantly advances the foundational concepts of the Capital Market Line (CML) and the Capital Asset Pricing Model (CAPM)~\cite{capm1, capm3} by incorporating unique investor perspectives, making it particularly valuable for managing portfolios with specific risk-return objectives.

Within our context, the BL model, adhering to its foundational premises, adjusts the weights of suppliers in future order allocations by incorporating historical perspectives that the enterprise holds towards its suppliers. The model encapsulates several critical components:
\begin{itemize}[leftmargin=*]
  \item \textbf{Equilibrium Supplier Profits ($\Pi$):} Represents the normalized expected returns of suppliers, based on the market equilibrium. We calculate $\Pi \in \mathbb{R}^{N}$ using $\mu$ as follows:
  \begin{equation}
  \label{eq: pi initiation}
    \Pi_i = {\exp{(\mu_i)}} / {\sum_j^N\exp{(\mu_j)}}.
    \end{equation}
where $\mu$ indicates the anticipated order returns from each supplier, highlighting the model's reliance on supplier market consensus as a starting point for adjustments based on enterprise insights. 
  \item \textbf{Perspective Matrix ($\mathcal{P}$):} This matrix $\mathcal{P}\in \mathbb{R}^{N\times N}$ converts the enterprise's subjective views into a framework of implied excess profits, where $\mathcal{P}_{ij}$ quantifies the supply capacity of supplier $\mathcal{SU}_{i}$ relative to $\mathcal{SU}_{j}$. A positive $\mathcal{P}_{ij}$ implies that $\mathcal{SU}_{i}$ is deemed more competitive than $\mathcal{SU}_{j}$, with the specific value indicating the degree; conversely, a negative value suggests the opposite. It is a critical component that allows the BL model to adjust the base equilibrium returns by incorporating the enterprise's specific insights or expectations about future supplier performance.
  \item \textbf{Perspective Return Vector ($\mathcal Q$):}  =
The vector $\mathcal{Q}\in \mathbb{R}^{N}$ that embodies the adjusted return rates expected from suppliers, factoring in the enterprise's perspectives. This vector is crucial for recalibrating the initial market-based profits ($\Pi$) to reflect the enterprise insights.
  \item \textbf{Error Covariance Matrix ($\Omega$):} The matrix $\Omega \in \mathbb{R}^{N\times N}$ encapsulates the uncertainty or confidence level in the enterprise's perspectives towards suppliers, as described by:
\begin{equation}
  \label{eq: pqmu}
\mathcal{P} \times \mu = \mathcal{Q} + N(0,\Omega).
\end{equation}
\end{itemize}

Leveraging Bayesian theory, we simplify $(\mathcal{O}-\mathcal{S})^\kappa$ to $\Sigma$, enabling a nuanced adjustment of the profit and risk components, as demonstrated by:
\begin{align}
    \label{eq: modified profit and risk item} 
    \begin{cases}
        & \widehat{\mu} = \Pi + \tau \Sigma \mathcal{P}^T (\mathcal{P}\tau\Sigma\mathcal{P}^T + \Omega)^{-1} (\mathcal{Q}-\mathcal{P}\Pi), \\
        & \widehat{\Sigma} = (1+\tau) \Sigma - \tau \Sigma \mathcal{P}^T (\mathcal{P} \tau \Sigma \mathcal{P}^T + \Omega)^{-1} \mathcal{P}\tau \Sigma,
    \end{cases}
\end{align}
where $\tau$ is a hyper-parameter that regulates the impact of the enterprise's perspective on the optimization outcome. The resultant matrices, $\widehat{\mu}$ and $\widehat{\Sigma}$, encapsulate the adjusted risk-return profile, guiding the allocation strategy towards the alignment that reflects the enterprise's expectations and market realities. Incorporating $\widehat{\mu}$ into the profit component and $\mathcal{W}^T\widehat{\Sigma}\mathcal{W}$ into the risk component yields the solution:
\begin{equation}
  \label{eq: BL solution}
\mathcal{W}^*_{BL} = (\delta \widehat{\Sigma})^{-1} \widehat{\mu},
\end{equation}
where $\mathcal{W}^*_{BL}$ denotes the optimal allocation strategy, with $\sum \mathcal{W}^*_{BL} = 1$, and accounts for the historical interactions with suppliers as well as the enterprise's perspective.
We harmonize historical supplier interactions with forward-looking perspectives, thereby offering a robust framework for strategic decision-making in supply chain management. \M's~ ability to systematically integrate and balance both market-derived equilibriums and specific enterprise insights distinguishes it as a powerful tool for portfolio optimization in a supply chain context.

\section{Algorithm Time Complexity}
\label{ap: algorithm complexity}
We analysis the time complexity of the whole \M~ as follows:
\begin{itemize}[leftmargin=*]
    \item The complexity of feature initiation and dynamic graph propagation matrix is $\mathcal{O}(NTp+NT) \approx \mathcal{O}(NTp)$ and $\mathcal{O}(N^2T+NT)\approx \mathcal{O}(N^2T)$, respectively. Once prepared, the data can be stored and reused during training and testing.
    \item In STGNNs Backbone Encoder, (i) Chebyshev polynomial convolution layer costs $\mathcal{O}(TC|E|ND^2p)$, where $|E|$ is the edge number and $D$ is the hidden dimension; (ii) Temporal Convolution Layer costs $\mathcal{O}(\mathcal{TN}D^2p)$; (iii) For the attention fusion layer, calculating $\mathcal{P}$ requires $\mathcal{O}(TNDp+ TN^2p + TN^2Dp) \approx \mathcal{O}(TN^2Dp)$ and $\Omega$ needs $O(TN^3p)$.
    \item For prediction, the solution of the BL model costs $\mathcal{O}(TN^3p)$, and
    the predictor needs $\mathcal{O}(TNDf)$.
\end{itemize}

\section{Robustness Study (RQ2)}
\label{ap: robutness}
\begin{figure}[htbp]
  \centering
  \hspace{-0.2cm}
\includegraphics[width=0.47\textwidth]{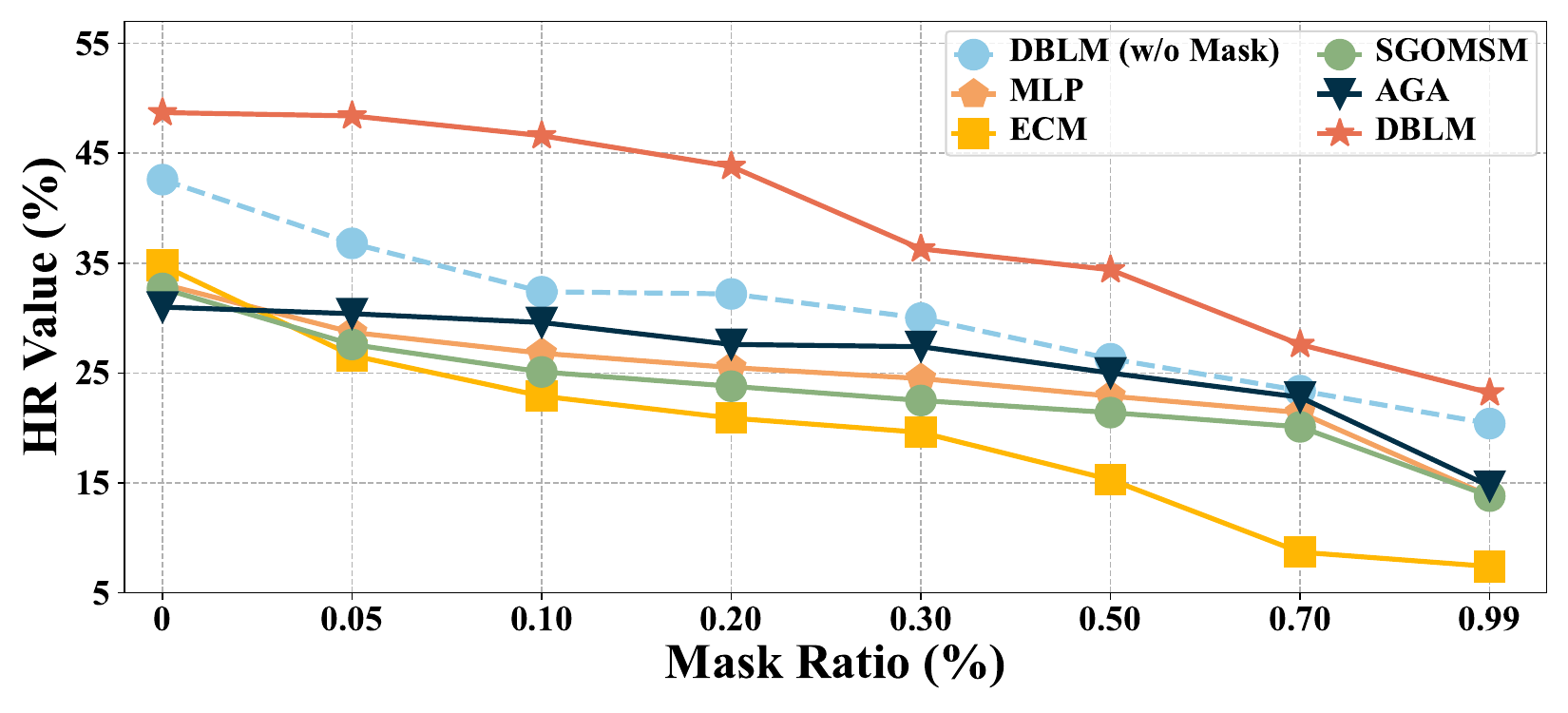}
  \caption{Robustness study of HR@50 by randomly masking $\mathcal{O}$ and $\mathcal{S}$ in training data on MCM dataset. Mask Ratio is from 0 to 0.99.}
  \label{fig:mask.PNG}
\end{figure}

To further answer RQ2, and to substantiate our model's proficiency in mitigating \textbf{\textit{C3. Data Unreliability}}, we artificially constructed unreliable data by implementing it through a masking mechanism on $\mathcal{S}$ and $\mathcal{O}$ randomly, with mask ratios varying from 0 to 1 on the MCM dataset. We report the HR@50 metric and the results are illustrated in Figure \ref{fig:mask.PNG}. Our observations are as follows:

\textbf{Trend Analysis: } The gradual increase in mask ratio typically tests the model's ability to maintain performance despite the progressive bias introduced by unreliable data.
\M~ exhibits superior resilience, evidenced by a slower decline in HR@50 compared to other models with increasing mask ratios, indicating its robustness to handle challenges posed by data unreliability.

\textbf{Performance Benchmark: } The \M~ model's performance, shows reduced sensitivity to the escalating mask ratios, highlighting its robustness in comparison to other baseline models. Notably, \M~ maintains a significantly higher HR@50 even at extreme mask ratios (e.g., 0.99), compared to other models (SGOMSM, MLP, AGA, ECM) would underscore its effectiveness in leveraging the remaining unmasked data.

It proves that \M~ is particularly well-adapted for scenarios with unreliable data, greatly mitigating the bias, and making it a potentially valuable approach for real-world TSSO applications. 

\section{Evaluation Matrics}
\label{sec: evaluation matrics}
\paragraph{\textbf{Hit Ratio}}
HR@K evaluates the model's accuracy in pinpointing the most critical risk points. It does this by assessing whether the largest $K$ ground-truth risk points are among the smallest $K$ predicted allocation weights. The computation of HR@K is as follows:
\begin{align}
\text{HR@K} = \frac{1}{\mathcal{B}_s f}\sum^{\mathcal{B}_s} \sum_{i=t}^{t+f} \frac{\# \text{TopK}(\Sigma_{i}) \cap \text{BottomK}(\hat{\mathcal{W}}^*_{i}) }{K},
\end{align}
where $\mathcal{B}_s$ denotes the batch size, and the calculation extends over the mean value of each predicted time horizon from $t$ to $t+f$. A higher HR@K ratio indicates a better capability of the model to identify the most critical risk points accurately.

\paragraph{\textbf{Mask Risk Expect}}
To complement our model evaluation and mitigate the impact of unreliable data, we introduce the MRE metric. This approach involves manipulating the Softmax function to set the allocation weights of unreliable suppliers to zero, while appropriately adjusting the remaining weights. The MRE is subsequently calculated over the time horizon from $t$ to $t+f$, by summing the products of the adjusted weights and their corresponding risk values. The equation for this calculation is detailed as follows:
\begin{align}
& \mathcal{W}^{(test)}_{i} = \text{Softmax}(\hat{\mathcal{W}}^*_{i} - \inf \odot \mathcal{M}_{i}), \\
& \text{MRE} = \frac{1}{\mathcal{B}_s f}{\sum^{\mathcal{B}_s}\sum_{i=t}^{t+f}    \Sigma_i^T  \times\mathcal{W}^{(test)}_{i}},
\end{align}
where $\mathcal{M}_i$ is the mask matrix to represent the timeslot $i$ unreliable suppliers. We first set the unreliable nodes' weight to zero and then calculate the batch mean MRE.
The lower the value of MRE, the higher the model's ability to avoid high-risk suppliers after processing untrustworthy data.

\section{Theoretical Analysis of the BL under Data Unreliability}
In this section, due to the challenge of \textit{\textbf{C3: Data Unreliability}}, which prevents the BL model from deriving an analytical solution. We delve into the underlying causes of this issue and present two viable solutions. In subsections, we involve a detailed analysis of the matrices' rank within each equation, supported by the application of the following theorems: 
\begin{theorem}
Let matrices \(A\), \(B\), and \(C\) be such that \(A = BC\), where \(B\) is an \(N \times F\) matrix, and \(C\) is an \(F \times N\) matrix. Then, the rank of matrix \(A\) satisfies the following condition:
\begin{align}
\text{Rank}(A) \leq \min\bigl(\text{Rank}(B), \text{Rank}(C)\bigl).
\end{align}
\end{theorem}

\begin{theorem}
\label{th 1}
Let \(A\) and \(B\) be matrices of the same dimensions. The rank of the sum of \(A\) and \(B\), denoted as \(\text{Rank}(A + B)\), satisfies the following inequalities:
\begin{align}
\text{Rank}(A + B) \leq \text{Rank}(A) + \text{Rank}(B)
\end{align}
\end{theorem}

\begin{theorem}
\label{th 3}
If \(A\) is an \(N \times N\) non-full-rank matrix and \(D\) is a full-rank \(N \times N\) diagonal matrix, then \(A + D\) is full-rank. The proven is as follows:

Given Rank(A) < N, Rank(D) = N, we can assume that  \begin{align*}
    &(A + D)x = 0, x \neq 0 \in \mathbb{R}^N  \\
&\text{Then} \quad Ax = -Dx.
\end{align*}
For D is invertible,  multiply both sides by \(D^{-1}\) : 
\begin{align}
    D^{-1}Ax = -x.
\end{align}
The transformation $D^{-1}A$ maps x to -x,   x $\neq$ 0 which implies $D^{-1}A  $ is full-rank. This contradicts to the fact that A is a non-full-rank matrix so A + D cannot be non-full-rank. Therefore A + D is full-rank.
\end{theorem}

\begin{lemma}
\label{lemma}
A square matrix \(A\) of size \(N \times N\) is invertible if and only if its rank is full, i.e., \(\text{Rank}(A) = N\). 
\end{lemma}

\subsection{Sigmoid Calibration}
Due to data unreliability, traditional computation of ${\Omega}$ complicates deriving the inverse of $(\mathcal{P}\tau\Sigma\mathcal{P}^T + \Omega)$, essential for the Black-Litterman (BL) model solution. To address this, we investigate the ranks of $\mathcal{P}$ and $\Omega$, incorporating sigmoid calibration for adjustment.
\label{sigmoid}
\subsubsection{Rank Analysis of $\mathcal{P}$}
The perspective matrix $\mathcal{P}$, an $N \times N$ square matrix, encapsulates the competitive dynamics among suppliers. To construct $\mathcal{P}$ leveraging both market (spatial) and historical (temporal) insights, we employ a Spatio-Temporal (ST) Attention mechanism. This approach utilizes spatial embeddings $\mathcal{E}\in \mathbb{R}^{N \times F}$ and temporal embeddings $\mathcal{H}\in \mathbb{R}^{N \times F}$, fusing them to produce $\mathcal{P}\in \mathbb{R}^{N \times N} \simeq  \mathbf{W}_{\text{attn}} \times \mathcal{H}\times\mathcal{E}$. However, if spatial and temporal embeddings are low-rank tensors, their combined rank could significantly fall below $N$, indicating a potential reduction in the information captured by $\mathcal{P}$:
\begin{align}
& \text{Rank}(\mathcal{H}) \leq F = \min(N, F)
\\
& \text{Rank}(\mathcal{E}) \leq F = \min(N, F).
\end{align}
As a result, leveraging Theorem \ref{th 1}, by deducing the case of multiplying spatial and temporal matrices with learnable matrix, the rank of the Perspective Matrix $\mathcal{P}$, as obtained through the Spatio-Temporal (ST) attention mechanism, can be determined as follows:
\begin{align}
\text{Rank}(\mathcal{P}) & \leq \min\bigl(\text{Rank}(\mathcal{H}), \text{Rank}(\mathcal{E}), \text{Rank}(\mathbf{W_{attn}}) \bigl)
\nonumber\\ & = \min \bigl(F, \text{Rank}(\mathbf{W_{attn}}) \bigl).
\end{align}
Thus, irrespective of whether $\text{Rank}(\mathbf{W_{attn}}) = N$ or $\text{Rank}(\mathbf{W_{attn}}) < N$, the matrix $\mathcal{P}$ does not achieve full rank status. Given that the rank of the learnable parameter $\mathbf{W}_{attn}$ is independent of the derivation, we simplify the expression by setting the rank of $\mathbf{W}{\text{attn}}$ equal to $F$, denoted as $\text{Rank}(\mathbf{W}_{\text{attn}})=F$. This simplification acknowledges that the rank constraints of $\mathcal{P}$ are primarily dictated by the dimensions of the spatial and temporal embeddings, and not directly by the dimension of the attention weights themselves.

\subsubsection{Rank Analysis of $\Omega_{old}$}
\label{ap: onega}
Traditional methods that do not leverage deep learning techniques compute $\Omega_{old}$ as follows:
\begin{align}
    \Omega_{old} = \text{diag}(\mathcal{P}\Sigma\mathcal{P}^T).
\end{align}
However, given the issue of data unreliability, $\Sigma$ manifests as a non-full rank diagonal matrix with $\text{Rank}(\Sigma)<N$, primarily due to the presence of zero risk elements. Since both $\mathcal{P}$ and $\Sigma$ are non-full rank, the rank of $\Omega_{old}$ is determined as:
\begin{align}
  \text{Rank}(\Omega_{old}) & \leq \min \bigl( \text{Rank}(\mathcal{P}), \text{Rank}({\Sigma}), \text{Rank}(\mathcal{P}^T)\bigl) \nonumber \\
  & = \min \bigl( F, \text{Rank}({\Sigma})\bigl) < N.
\end{align}
Therefore, $\Omega_{old}$ also emerges as a non-full rank matrix. This rank limitation becomes particularly problematic when adjusting $\mu$ and $\Sigma$, as it necessitates the inverse operation $\mathcal{P}\tau\Sigma\mathcal{P}^T + \Omega_{old}$, denoted as $\mathcal{J}_{\text{old}}$, which is unattainable due to the rank deficiency.

\subsubsection{Unachievement of $\mathcal{J}_{\text{old}}$}
\label{ap: unj}
Given Lemma \ref{lemma} stipulates that a matrix must be full rank to be invertible, the rank of $\mathcal{J}_{\text{old}}$ is crucial for the reversibility required in solving the BL model. We analyze the rank of $\mathcal{J}_{\text{old}}$ as follows:
\begin{align}
\text{Rank}(\mathcal{J}_{\text{old}}) & \leq  \text{Rank}(\mathcal{P}\tau\Sigma\mathcal{P}^T)+ \text{Rank}(\Omega_{old})
\nonumber \\ \nonumber
& = \min \bigl(F, \text{Rank}(\mathbf{W_{attn}}), \text{Rank}(\Sigma) \bigl) + \text{Rank}(\Omega_{old}) 
\\ \nonumber &
\leq \min \bigl(F, \text{Rank}(\Sigma) \bigl) + \min \bigl(F, \text{Rank}(\Sigma) \bigl) \\
& = 2 \min \bigl(F, \text{Rank}(\Sigma) \bigl).
\end{align}
Thus, when $\min \bigl(F, \text{Rank}(\Sigma)\bigl) < N/2$, it implies that $\mathcal{P}\tau\Sigma\mathcal{P}^T + \Omega_{\text{old}}$ does not satisfy the full rank condition, making it non-invertible. This rank deficiency due to data unreliability hampers the computation of $\mathcal{W}^*$, highlighting the challenges in applying the traditional BL model formulation under these constraints.

\subsubsection{Solution to transfer $\Omega_{\text{old}}$}
\label{ap: sigmoid_1_sol}
To calibrate $\mathcal{J}_{\text{old}}$ for better computation, we refine the approach for calculating $\Omega$ by introducing learnable parameters $\mathbf{W}_{\text{om}}$ and $\mathbf{B}_{\text{om}}$, as outlined in Eq \eqref{eq:omega}. Despite these parameters facilitating linear transformations, they inherently do not augment the rank of $\Omega_{\text{old}}$. To circumvent this limitation, we incorporate a nonlinear activation function, specifically the Sigmoid function $\sigma(\cdot)$, which is applied element-wise. This choice is strategic; unlike ReLU, ELU, Tanh, and SeLU, the Sigmoid function ensures that no zero values are output after its application. Thus, when diagonal elements are extracted post-Sigmoid application, $\Omega$ attains full rank status: $\text{Rank}(\Omega) = N$.

As highlighted in Theorem \ref{th 3}, combining a non-full-rank matrix with a full-rank diagonal matrix results in a full-rank matrix. Notably, $\Omega$, being a diagonal matrix, achieves a rank of $N$. This is significant because even though $\mathcal{P}\tau\Sigma\mathcal{P}^T$ may not be full-rank, the resultant matrix $\mathcal{J}$ achieves full-rank status.

Consequently, $\mathcal{J}$ becomes invertible. The introduction of the Sigmoid function resolves the rank deficiency issue, ensuring the Black-Litterman model's applicability under conditions of data unreliability.

\subsection{Regularization Calibration}
\label{ap: rc}
In solving Eq \eqref{eq: BL solution}, it's critical that $\widehat{\Sigma}$ ought to be full rank to enable an invertible transformation. We revisit $\widehat{\Sigma}$ for a more detailed analysis:
\begin{align}
    \label{eq: simplified widehat_sigma}
        & \widehat{\Sigma} = \Sigma \left((1+\tau) I - \tau \mathcal{P}^T \left(\mathcal{P} \tau \Sigma \mathcal{P}^T + \Omega\right)^{-1} \mathcal{P}\tau \Sigma\right),
\end{align}
where \(I\) represents the identity matrix. We then analyzing the rank of $\widehat{\Sigma}$ as:
\begin{equation}
\begin{aligned}
    \label{eq: rank 2 a}
         \text{Rank}({\widehat{\Sigma}}) 
        & \leq \min \Bigl(\text{Rank}(\Sigma), \text{Rank}\bigl( (1+\tau) I - \tau \mathcal{P}^T \\ 
        & \times (\mathcal{P} \tau \Sigma \mathcal{P}^T + \Omega)^{-1} \mathcal{P}\tau \Sigma\bigl) \Bigl) \\
        & = \min \bigl(\text{Rank}(\Sigma), N\bigl) < N.
\end{aligned}
\end{equation}
Interestingly, $\text{Rank}\bigl((1+\tau) I - \tau \mathcal{P}^T (\mathcal{P} \tau \Sigma \mathcal{P}^T + \Omega)^{-1} \mathcal{P}\tau \Sigma\bigl) = N$ by Theorem ~\ref{th 3}, with $(1+\tau) I$ acting as the full-rank diagonal matrix and $\text{Rank}\bigl(\tau \mathcal{P}^T (\mathcal{P} \tau \Sigma \mathcal{P}^T + \Omega)^{-1} \mathcal{P}\bigl)\leq \min \big(\text{Rank}(\mathcal{P}), \text{Rank}(\Sigma), \text{Rank}({\Omega}) \bigl) \\ = \min \big(\text{Rank}(\mathcal{P}), \text{Rank}(\Sigma), N) \bigl) < N$, acting as non-full-rank matrix.
Thus, the rank of $\widehat{\Sigma}$ hinges solely on the first term. To address this and counter non-invertibility, we apply a regularization calibration by adding a small term, $\epsilon I$, where $\epsilon$ is a minuscule positive constant (e.g., $1\times10^{-4}$). This adjustment guarantees that $\widehat{\Sigma}$ attains full rank and becomes invertible, as supported by Theorem ~\ref{th 3}:
\begin{equation}
\small
\begin{aligned}
    \label{eq: rank 2 adjust}
         \text{Rank}({\widehat{\Sigma}}) 
       & = \text{Rank}\left((\Sigma+\epsilon I)\bigl((1+\tau) I - \tau \mathcal{P}^T (\mathcal{P} \tau \Sigma \mathcal{P}^T + \Omega)^{-1} \mathcal{P}\tau \Sigma \bigl)\right) \\
        & = \text{Rank}(\Sigma+\epsilon I) = N.
\end{aligned}
\end{equation}
The regularization calibration is then denoted as follows:
\begin{equation}
 \widehat{\Sigma} = (1+\tau) (\Sigma +\epsilon I) - \tau (\Sigma+\epsilon I) \mathcal{P}^T \times \bigl(\mathcal{P}  \tau \Sigma  \mathcal{P} ^T  
         +\Omega \bigl)^{-1} \mathcal{P} \tau \Sigma.
\end{equation}




\section{Baselines Details}
\label{ap: baselines details}
\begin{itemize}[leftmargin=*]
\item \textbf{HA}~\cite{10.1609/aaai.v33i01.3301922}: Historical Average method, which calculates future allocation weights by averaging the allocation weights of the previous $p$ time slices. This approach provides a straightforward prediction based on historical trends.
\item \textbf{MC}~\cite{Puka2022FuzzyMA}: The Monte Carlo simulation employs random sampling and probabilistic models to analyze complex systems. For our application, we generate 10,000 random weight combinations (ensuring each sample's weight sum equal to 1) and identify the combination that minimizes the Mean Relative Error (MRE) over the last $p$ periods as the forecast for the next $f$ periods.
\item \textbf{Greedy}~\cite{9835149}: This method adopts a step-wise optimization strategy, prioritizing immediate gains. It iteratively allocates zero weight to the highest-risk option and the maximum weight to the lowest-risk option, with a subsequent normalization step to ensure the sum of weights equals 1. This process aims to exploit historical data to guide future allocations.
\item \textbf{DP}~\cite{Kuroiwa_2023}: Dynamic programming addresses complex decision-making by decomposing it into simpler subproblems. Starting with equal weights, the algorithm iteratively adjusts the allocation weights to minimize risk based on historical performance.
\item \textbf{Fuzzy-AHP}~\cite{rezaei2020supplier}: A multi-criteria decision-making method that integrates fuzzy logic with the Analytic Hierarchy Process (AHP) to tackle uncertainty and ambiguity issues, assisting decision-makers in making trade-offs and choices in complex environments. 
In our application, supplier weights are determined based on historical data and a judgment matrix ($JM$). The $JM$ is meticulously crafted through a comparative analysis of the significance of eight distinct features: $f^{sv}_{it}, f^{ov}_{it}, f^{sr}_{it}, f^{ssv}_{it}, f^{hsr}_{it}, f^{hssv}_{it}, f^{sc}_{it}, f^{ss}_{it}$: 
\begin{align}
JM = 
\begin{pmatrix}
    1 & 2 & 1.333 & 4 & 0.8 & 2 & 0.571 & 1.333 \\
0.5 & 1 & 0.667 & 2 & 0.4 & 1 & 0.286 & 0.667 \\
0.75 & 1.5 & 1 & 3 & 0.6 & 1.5 & 0.429 & 1 \\
0.25 & 0.5 & 0.333 & 1 & 0.2 & 0.5 & 0.143 & 0.333 \\
1.25 & 2.5 & 1.667 & 5 & 1 & 2.5 & 0.714 & 1.667 \\
0.5 & 1 & 0.667 & 2 & 0.4 & 1 & 0.286 & 0.667 \\
1.75 & 3.5 & 2.333 & 7 & 1.4 & 3.5 & 1 & 2.333 \\
0.75 & 1.5 & 1 & 3 & 0.6 & 1.5 & 0.429 & 1 \\
\end{pmatrix}.
\end{align}
\item \textbf{Fuzzy-TOPSIS}~\cite{hasan2020resilient}: The Fuzzy-TOPSIS method is a multi-criteria decision-making approach that determines the weights (importance) of suppliers by handling fuzzy data, thus considering uncertainty in supplier selection and order allocation issues. Based on history data, we first positive process these 8 features $f^{sv}_{it}, f^{ov}_{it}, f^{sr}_{it}, f^{ssv}_{it}, f^{hsr}_{it}, f^{hssv}_{it},  f^{sc}_{it}, f^{ss}_{it}$. Then we use the TOPSIS method to calculate the allocation weight.
\item \textbf{Markowitz}~\cite{way2018wright}: The Markowitz algorithm, optimizes portfolio allocation by balancing return and risk through diversification based on the covariance of asset returns. Similar to \M, we build the risk item and the profit item based on historical data, and then we use the solution of $\mathcal{W}^*$ to predict the future $f$ time period.
\item \textbf{DT}~\cite{Bowser_Chao_1993}: Decision Trees map out decisions, uncertainties, and outcomes, offering a visual exploration of decision-making pathways. Our adaptation regresses future $1/\Sigma$ values, normalizing these predictions to formulate allocation weights.
\item \textbf{Lasso}~\cite{Tibshirani1996}: Similar to DT, we use the Lasso regressor to predict the future $1/\Sigma$ and normalized the prediction as allocation weight.
\item \textbf{MLP}~\cite{MLP}: MLP simply utilizes the multi-layer perception to perform regression tasks. We move forward to leverage Mask Spearman Rank Loss to supervise model training.  
\item \textbf{ECM}~\cite{Xing_Cambria_Malandri_Vercellis_2018}: ECM introduces a method for optimizing asset allocation with a Bayesian model via an LSTM encoder.  
\item \textbf{SGOMSM}~\cite{hui2023constrained}: SGOMSM tackles the complex issue of adjusting allocation weight by introducing a two-stage optimization method that combines high-level guidance with detailed modeling of supplier correlations. 
\item \textbf{AGA}~\cite{li2021large}: AGA introduces a framework that leverages neural networks for accurate market predictions and an innovative adaptive gradient ascent (AGA) method for efficient optimization.
\end{itemize}

\end{document}